\documentclass{article}
\usepackage{parskip}
\usepackage[margin=1.0in]{geometry}

\usepackage{graphicx}
\usepackage{caption}
\usepackage{subcaption}
\usepackage{tikz}
\usepackage{float}

\usepackage{enumitem}

\usepackage{physics}
\usepackage{amsmath}
\usepackage{amsthm}
\usepackage{amstext}
\usepackage{amssymb}
\usepackage{thmtools}
\usepackage{siunitx}
\usepackage{bm}
\theoremstyle{definition}

\newtheorem{exmp}{Example}[]

\usepackage[english]{babel}

\usepackage{pdfpages}

\usepackage{dsfont}
\usepackage{xcolor}
\definecolor{amber}{rgb}{1.0, 0.6, 0.0}
\usepackage{textcomp}
\usepackage{listings}
\usepackage{relsize}
\usepackage{multicol}
\usepackage{hyperref}
\hypersetup{
	colorlinks=true,
	linkcolor=black,
	filecolor=black,      
	urlcolor=black,
	citecolor=blue,
}
\usepackage[capitalise]{cleveref}
\crefname{figure}{Fig.}{Fig.}
\crefname{table}{Table}{Tables}
\crefname{equation}{Eqn.}{Eqns.}
\crefname{algocf}{Algorithm}{Algorithms}
\crefname{exmp}{Example}{Ex.}

\crefname{lemma}{Lemma}{Lemmas}
\crefname{corollary}{Corollary}{Cor.}
\crefname{defn}{Definition}{Definitions}

\usepackage{physics}
\usepackage{url}

\usepackage{pifont}

\usepackage{bbm}
\usepackage{mathtools}
\usepackage{hhline}
\usepackage{tabularx}

\newcommand{\R}{\mathbb{R}}

\newcommand{\mcU}{\mathcal{U}}
\newcommand{\mcV}{\mathcal{V}}
\newcommand{\mcC}{\mathcal{C}}
\newcommand{\mcL}{\mathcal{L}}

\newcommand{\mfe}{\mathfrak{e}}
\newcommand{\mfd}{\mathfrak{d}}

\usepackage[ruled,vlined]{algorithm2e}
\SetKwInOut{KwParams}{Parameters}

\usepackage{authblk}

\title{Conformal Disentanglement and Latent-Space Curation:\\A Neural Framework for Perspective Synthesis, Differentiation \\ and Targeted Generation}
\author[1,3]{George A. Kevrekidis}
\author[2]{Eleni D. Koronaki}
\author[1, 4]{Dimitris G. Giovanis}
\author[1,5]{\\Ioannis G. Kevrekidis}

\affil[1]{Department of Applied Mathematics and Statistics, Johns Hopkins University, Baltimore, MD, USA}
\affil[2]{Faculty of Science, Technology and Medicine, University of Luxembourg, Esch-sur-Alzette, Luxembourg}
\affil[3]{Los Alamos National Laboratory, Los Alamos, NM, USA}
\affil[4]{Department of Civil and Systems Engineering, Johns Hopkins University, Baltimore, MD, USA}
\affil[5]{Department of Chemical and Biomolecular Engineering, Johns Hopkins University, Baltimore, MD, USA}

\date{June 2026 \\ LA-UR-24-28862}

\begin{document}

\maketitle

\begin{abstract}
Many scientific and engineering problems involve observing a common phenomenon through multiple heterogeneous sensors or measurement modalities. Such observations typically contain both information {\it shared} across sensors, reflecting the underlying system, and {\em sensor-specific or extraneous} components arising from measurement processes or environmental effects. Disentangling these contributions is essential when sensor-independent observations are unavailable. We propose a neural autoencoder framework that explicitly separates shared and sensor-specific latent variables from multi-sensor data. The architecture enforces geometric independence between latent components through structural constraints and orthogonality-based regularization, yielding interpretable and disentangled representations. Building on this representation, we then introduce {\em a latent-space generative methodology} in which generative models are tuned/``restricted" on selected disentangled latent subspaces; we then constructively combine disentangled observed latent variables to conditionally synthesize new samples via trained decoders. This enables consistent data generation with prescribed shared (or sensor-specific) characteristics. It also supports cross-sensor inference by consistently sampling distributions over plausible measurements in unobserved modalities. We demonstrate the approach on several computational examples, showing effective disentanglement, targeted data generation, and modality imputation in heterogeneous sensing settings.
\end{abstract}

\section{Introduction}

In many scientific and engineering applications, a system of interest is observed through multiple heterogeneous sensors or measurement modalities. These sensors may differ in targeted physical observables, viewpoints, spatial or temporal resolution, or measurement mechanism - and therefore provide complementary yet entangled descriptions of the same underlying phenomenon. As a result, each observation typically reflects a combination of latent information that is shared across sensors, sensor-specific components arising from instrumental effects, environmental interactions, or auxiliary dynamics. Identifying what is common across observations, while disentangling what is specific to each sensor, constitutes a fundamental challenge in multi-sensor data analysis.

This class of problems has been studied under the complementary notions of Perspective Synthesis and Perspective Differentiation \cite{Ekeroth2023thesis}, which we will collectively refer to as the ``Common \& Uncommon Variable" problem. Perspective Synthesis concerns the identification of latent variables that are shared across multiple observers, whereas Perspective Differentiation addresses the separation of sensor-specific components that remain once the common structure has been identified. The latter task is only meaningful after the former has been resolved, and is closely related to the broader notion of disentanglement in representation learning.

An intuitive illustration of this setting, originally introduced in \cite{lederman2014,lederman2018}, is the following.

\begin{figure}
\centering
\includegraphics[width=\linewidth]{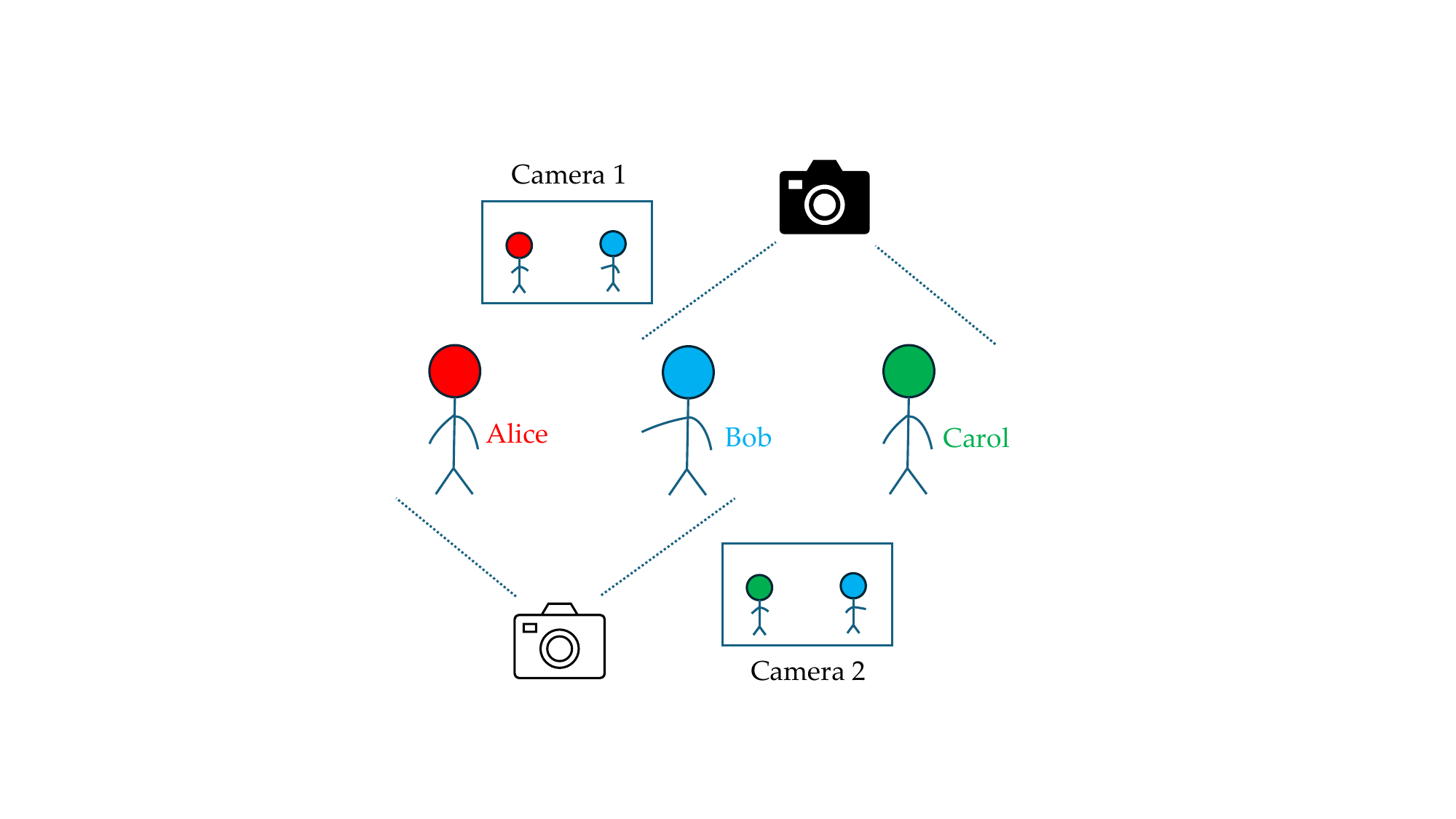}
\caption{Caricature described in \cref{exmp:intro}. Bob (blue, center) is observed from two different perspectives by two cameras. Alice (red, left) is observed only by Camera~1, and Carol (green, right) only by Camera~2.}
\label{fig:example}
\end{figure}

\begin{exmp}\label{exmp:intro}
Three individuals (Alice, Bob, and Carol) stand in a row and move (in this case, spin with distinct -not locked- frequencies) over time, while two cameras record their motion (\cref{fig:example}). Camera~1 observes Alice and Bob, whereas Camera~2 observes Bob and Carol. Both cameras capture Bob, but from different viewpoints and without aligned perspectives. 
Given long, simultaneously recorded sequences from both cameras, the objective is to intrinsically identify Bob as a common underlying entity across sensors, while Alice and Carol are distinct entities. Moreover, given Bob’s observation from one camera at a given time, we seek to infer the corresponding observation from the other camera. In systems terminology, this amounts to constructing an observer that maps observations from one sensor to consistent possible observations of the other.
\end{exmp}

In this and more general settings, the common variables correspond to latent degrees of freedom that are invariant across sensing modalities, while the uncommon variables encode sensor-specific information. In realistic scenarios, however, these components are rarely separable at the level of raw observations. Unlike idealized cases in which each measurement channel is influenced by only a single latent factor (a “clean” observation setting), practical sensing systems often produce measurements that depend {\em jointly on both} common and uncommon variables (a “dirty” observation setting). This coupling complicates inference across sensors and obscures what one sensor’s observation implies about another’s.

A range of classical and modern approaches have addressed aspects of this problem. Kernel-based spectral methods, such as Diffusion Maps \cite{COIFMAN20065}, and in particular the Alternating Diffusion algorithm \cite{lederman2014,lederman2018}, have been proposed to extract common variables from multi-observer data. Variants of Output-Informed Diffusion Maps \cite{lafon2004diffusion,HOLIDAY2019419}, Jointly Smooth Functions \cite{params}, and related constructions \cite{coifman2023common} address the identification of sensor-specific structure. Connections also exist to Canonical Correlation Analysis \cite{hardle2015canonical}, nonlinear Independent Component Analysis \cite{hyvarinen1999ICA}, and contrastive learning approaches \cite{cherti2023reproducible}. While these methods provide valuable insights, many are limited in their ability to simultaneously disentangle shared and sensor-specific structure in highly nonlinear, high-dimensional, or generative settings.

In parallel, there is an increasing demand not only to analyze multi-sensor data, but also to generate it. Scientific modeling, simulation, and data augmentation workflows increasingly rely on the ability to synthesize plausible observations, infer missing modalities, or explore counterfactual scenarios—particularly in settings with limited or incomplete sensor coverage ~\cite{tashiro2021csdi,scholkopf2021toward}. These tasks require representations that are not only disentangled, but also amenable to controlled manipulation and generative sampling ~\cite{song2021scorebased, lecun2022path, assran2023selfsupervised}. In particular, the Joint Embedding Predictive Architecture (JEPA) framework introduced by LeCun and collaborators emphasizes predictive latent representations that capture controllable structure and causal dependencies without requiring full generative reconstruction.

To address this need, we incorporate \emph{score-based diffusion models}~\cite{song2020score} and \emph{probabilistic generative modeling on manifolds}~\cite{soize2016data} within a structured latent space formulation. These tools enable sampling from learned latent distributions in a manner that respects both the geometry and the semantic organization of the data. Generated latent embeddings —corresponding either to plausible unseen observations or to controlled perturbations of existing samples—are decoded back into the ambient space using trained decoders. This construction supports a range of capabilities, including simulating new sensor observations that preserve shared structure, varying sensor-specific effects while holding common content fixed, and estimating what one sensor could be measuring given observations from another. Such \emph{cross-sensor generative inference} can be interpreted as a learned translation between sensing modalities, closely related to the construction of level sets of compatible measurements, or to the synchronization of dynamical systems from partial observations; this is discussed in a previous (2024) version of the present manuscript, accessible through arXiv history feature. Similar inference problems arise in sensor-limited industrial contexts, where sparse or noisy measurements must support reliable reconstruction of unobserved states \cite{koronaki2023partial}.

The ability to control latent variations has further implications for downstream tasks. In low-data regimes, synthetic augmentation can improve robustness. In privacy-sensitive settings, sensor-specific components may be suppressed during generation. More broadly, the learned latent geometry reveals global structure across sensing systems, enabling tasks such as \emph{sensor fusion}, \emph{domain adaptation}, and \emph{modality translation}. Our work builds on prior research in \emph{common variable learning}~\cite{lederman2014common} and \emph{multi-sensor spectral analysis}~\cite{coifman2006diffusion, mishne2019diffusion}, while extending these ideas to a unified, generative framework.

\paragraph{Structure and Contributions} We propose and implement a unified approach that addresses both disentanglement and generation. First, we introduce a structured neural autoencoder architecture that explicitly separates common and sensor-specific latent variables from multi-sensor observations. The architecture is grounded in a geometric interpretation of disentanglement, viewing latent components as lying on orthogonal submanifolds, and enforcing block-orthogonality constraints to promote geometric independence. This design yields interpretable representations and consistent alignment across sensors.

Second, building on the learned disentangled representation, we extend the framework to a generative setting. Probabilistic models are learned directly on selected latent subspaces  —common, uncommon, or both— and used to synthesize new latent samples that are subsequently decoded into the ambient space. This enables controlled generation of observations with prescribed shared or sensor-specific characteristics; it naturally supports cross-sensor generative inference by producing distributions conditioned over compatible measurements across modalities.

The resulting bi-level framework couples disentanglement with latent-space generative control. We demonstrate its effectiveness on synthetic dynamical systems and high-dimensional image data, highlighting its utility for representation learning, targeted data generation, modality imputation, and multi-sensor inference. Beyond sensing applications, the proposed methodology provides a general approach to structured multi-view learning and --more generally-- to the data-driven modeling of complex systems.

\section{Problem Statement}\label{sec:Problem_Statement}
Mnemonically, we will use the upper- and lower-case letter `C' to label concepts related to the `common' system, and the upper- and lower-case letters `U' and `V' for concepts related to each of the `uncommon', sensor-specific systems. We describe the setting for two sensors observing a total of three systems, one of which is common; it is straightforward to extend to different settings with multiple sensors (involving arbitrary, possibly intricate choices of what is common and uncommon between different subsets of sensors).

\textbf{Setup:} We let
\begin{gather}\label{eqn:manifolds}
	\begin{split}
		\mcU\hookrightarrow \R^{n_u}\qc\dim\mcU=d_u\\
		\mcV\hookrightarrow \R^{n_v}\qc\dim\mcV=d_v\\
		\mcC\hookrightarrow\R^{n_c}\qc\dim\mcC=d_c
	\end{split}
\end{gather}
be three embedded submanifolds of Euclidean space where, in each case, the ambient ($n_u,n_v,n_c$) and intrinsic ($d_u,d_v,d_c$) dimensions are known. Here, $\mcC,\mcU,\mcV$ represent the common and two uncommon systems respectively. For each submanifold, it is possible for the ambient and intrinsic dimensions to be equal, but they generically are not; In the case where the common system $\mcC$ is a circle, for example, we would have $n_c=2$ and $d_c=1$. We will assume that all $n_u,n_v,n_c,d_u,d_v,d_c$ are known, and that the embedding Euclidean dimensions $n_u,n_v,n_c$ are minimal for smooth embeddings to exist \cite{whitney1992collected}.

Suppose that we are given two sets $S_1,S_2$ of sampled snapshots as data sampled simultaneously from the two sensors:
\begin{align}\label{eqn:samples}
\begin{split}
    S_1 = \qty{s_{u,i}\in\R^{k_u}}_{i=1}^N\\
    S_2 = \qty{s_{v,i}\in\R^{k_v}}_{i=1}^N
\end{split}
\end{align}
where the observed data dimensions $k_u$ and $k_v$ are possibly large. In particular, these will satisfy the following inequalities such that there is no information lost:
\begin{equation}\label{eqn:ambient_inequality}
    \begin{split}
        k_u\geq n_u+n_c\\
        k_v\geq n_v+n_c
    \end{split}
\end{equation}
though an additional dimension reduction step can convert the problem to one where equality holds in \cref{eqn:ambient_inequality}.

Furthermore, we assume that there exist two smooth, left-invertible maps ($\Phi_u,\Phi_v$) such that
\begin{equation}\label{eqn:data}
        \begin{split}
	   \Phi_u(S_1) = D_u= \qty{\vb{d}_{1,i}\in\mcU\times\mcC}_{i=1}^{N}\subset\R^{n_u+n_c}\\
	   \Phi_v(S_2) = D_v= \qty{\vb{d}_{2,i}\in\mcV\times\mcC}_{i=1}^{N}\subset\R^{n_v+n_c}
	\end{split}
\end{equation}
where individual snapshots can be written in their coordinate form
\begin{align}\label{eqn:embedding_coordinates}
	\begin{split}
		\vb{d}_1 = (u_1,...,u_{n_u},c_1,...,c_{n_c})\\
		\vb{d}_2 = (v_1,...,v_{n_v},c_1',...,c_{n_c}')
	\end{split}
\end{align}
which are coordinates of the respective Euclidean embedding spaces, restricted to the (non-linear) submanifolds. That is to say, even though the observations may be possibly high-dimensional ($k_u,k_v$), the data lie on lower-dimensional submanifolds in Euclidean space, and furthermore these \textbf{are `factorizable' into a product of two `disentangled' submanifolds, respectively, for \textit{each}  sensor}.

After applying $\Phi_u, \Phi_v$, we can define the natural projections $\pi_u,\pi_v,\pi_c,\pi_{c'}$ on the corresponding coordinates of the three disentangled submanifolds. We will additionally assume that there exists a diffeomorphism $\varphi$ on the common submanifold:
\begin{align}\label{eqn:diffeo}
	\varphi:\mcC\to\mcC\text{ such that } \varphi\circ\pi_c(\vb{d}_{1,i})=\pi_{c'}(\vb{d}_{2,i})
\end{align}
which establishes pointwise correspondence, meaning that the $\mcC$ appearing in the two data sets is `the same object'. Thus, $S_1, S_2$ can be shuffled by a permutation in their indices, as long as it is the same permutation across both data sets. The pair-correspondence \textit{between simultaneous snapshots} is what determines the common submanifold uniquely up to the diffeomorphism $\varphi$.

\textbf{Interpretation:} The setup generalizes the task described in the introduction: two sensors ($S_1,S_2$) make simultaneous observations, parts of which are (different) observations of the same `common' system. This is abstractly represented by the common submanifold $\mcC$. However, the common observations need not be identical, as long as they are equivalent up to a diffeomorphism $\varphi$ (similar to having two cameras observe the same person from different perspectives). Each sensor, additionally, also observes some amount of `uncommon' information which is `irrelevant' with respect to $\mcC$, represented respectively as the uncommon submanifolds $\mcU$ and  $\mcV$. It is therefore reasonable to ask whether, given only sensor data in the form of \cref{eqn:samples}, we can recover `intrinsic' parametrizations of {\em both} the common and each of the two uncommon systems.

\textbf{Objective:} Given samples of the form of \cref{eqn:samples}, we would like to: \textbf{(a)} Identify and parametrize the `common' submanifold $\mcC$; \textbf{(b)} Identify and parametrize each uncommon submanifold $\mcU,\mcV$. For now, we assume that topological and geometric characteristics of each submanifold are unknown, and we want to learn a representation of each submanifold as embedded individually in Euclidean space (\cref{eqn:manifolds}).

Ultimately, we also want to \textbf{(c)} {\em establish relations between observations of one sensor and (level sets of) consistent observations of the other sensor}.

\section{Methods}\label{sec:methods}
In this section we outline an approach to solving the problem described in \cref{sec:Problem_Statement} using bespoke autoencoder neural networks. Specifically, we develop an architecture and implement appropriate optimization algorithms, which are suitable for recovering the desired parametrizations of the common and uncommon submanifolds.

\subsection{Autoencoder Network Architectures}\label{sec:architecture}

\begin{figure}[ht]
	\centering
	\includegraphics{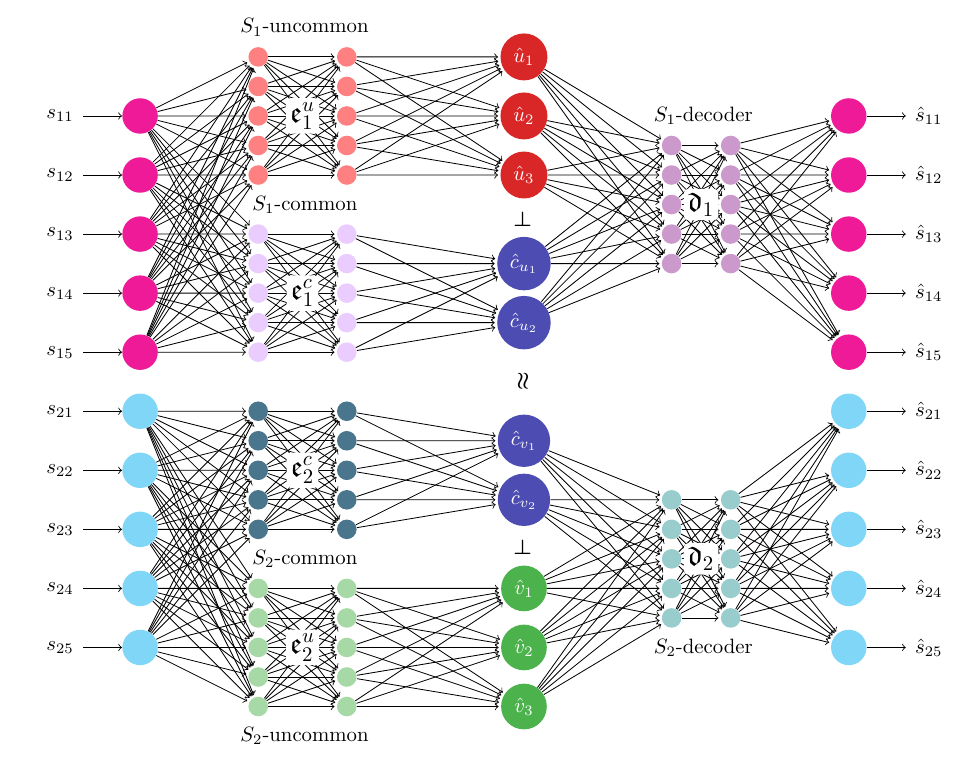}
	\caption{Sketch of the proposed network architecture outlined in \cref{sec:architecture}. The dimensions of the input, output and latent spaces correspond to those of \cref{exmp:RL}, while the encoder and decoder networks are just representations (and may have arbitrary width and depth)}.
	\label{fig:architecture}
\end{figure}

We define encoder-decoder pairs separately for each data set. Each encoder consists of common and uncommon parts as follows (see also \cref{fig:architecture}):

For $S_1$:
\begin{itemize}
	\item The encoder $\mfe_1:\R^{k_u}\to\R^{d_u+d_c}$ consists of two maps aimed at separating the common and uncommon parts:
	\begin{align}
		&\mfe_1^c:\R^{k_u}\to\R^{d_c}\qc\vb{s}_u\mapsto\hat{\vb{c}}_u\qq{(common)}\\ 
		&\mfe_1^u:\R^{k_u}\to\R^{d_u}\qc\vb{s}_u\mapsto\hat{\vb{u}}\qq{(uncommon 1)}
	\end{align}
	\item The decoder $\mfd_1:\R^{d_u+d_c}\to\R^{k_u},(\hat{\vb{c}}_u,\hat{\vb{u}})\mapsto\hat{\vb{s}}_u$ consists of a single map intended to approximate the left inverse of the entire encoder $\mfe_1$.
\end{itemize}
Similarly, for $S_2$:
\begin{itemize}
	\item The encoder $\mfe_1:\R^{k_v}\to\R^{d_u+d_c}$ consists of two maps aimed at separating the common and uncommon parts:
	\begin{align}
		&\mfe_2^c:\R^{k_v}\to\R^{d_c}\qc\vb{s}_v\mapsto\hat{\vb{c}}_v\qq{(common)}\\ 
		&\mfe_2^u:\R^{k_v}\to\R^{d_u}\qc\vb{s}_v\mapsto\hat{\vb{v}}\qq{(uncommon 2)}
	\end{align}
	\item The decoder $\mfd_2:\R^{d_v+d_c}\to\R^{k_v},(\hat{\vb{c}}_v,\hat{\vb{v}})\mapsto\hat{\vb{s}}_v$ consists of a single map intended to approximate the left inverse of the entire encoder $\mfe_2$.
\end{itemize}
A diagram of the proposed architecture is presented in \cref{fig:architecture}. In practice, each map is parametrized by a single fully connected multilayer  neural network with smooth activations.

\subsection{Bi-Level Optimization Algorithm}\label{sec:opti}
The goal of our optimization task is to approximate the maps $\Phi_u,\Phi_v$ and the embedding coordinates of \cref{eqn:embedding_coordinates}, up to diffeomorphisms. Upon successful training of the architecture parameters, the decoder outputs will be `good' reconstructions of the input data, i.e. each combined encoder ($\mfe_1,\mfe_2$) has a left inverse that is well-approximated by each of the decoders ($\mfd_1,\mfd_2$)
in some appropriate norm (usually specified to be $\ell^2$ (MSE) for individual points in the training data set).

There is a total of three objectives we aim to simultaneously satisfy:
\begin{align}
    \mcL_\text{reconstruction}&=\frac{1}{N}\sum_{i=1}^N\qty(\norm{\vb{s}_{u,i}-\hat{\vb{s}}_{u,i}}_2^2+\norm{\vb{s}_{v,i}-\hat{\vb{s}}_{v,i}}_2^2)\\
    \mcL_\text{common}&=\frac{1}{N}\sum_{i=1}^N\norm{\hat{\vb{c}}_{u,i}-\hat{\vb{c}}_{v,i}}_2^2\\
\mcL_\text{orthogonality}&=\frac{1}{N}\sum_{i=1}^N\qty(\sum_{j,k}\expval{\grad(\mfe_1^u(\vb{s}_{u,i}))_j,\grad(\mfe_1^c(\vb{s}_{u,i}))_k}+\sum_{l,k}\expval{\grad(\mfe_2^u(\vb{s}_{v,i}))_k,\grad(\mfe_2^c(\vb{s}_{v,i}))_l})
\end{align}
Individually, minimizing $\mcL_\text{reconstruction}$ ensures that each encoder is left-invertible, while minimizing $\mcL_\text{common}$ biases the common encoders ($\mfe_1^c,\mfe_2^c$) to find the same common latent embedding. Finally, $\mcL_\text{orthogonality}$ is a measure of (block-) functional independence, forcing the gradients of the functions parametrizing the uncommon systems within a sensor to be pointwise perpendicular to those parametrizing the common. Here, the indices $j,l$ range over $n_u,n_v$ respectively, while the index $k$ ranges over $n_c$. The inner product is inherited by the Euclidean ambient space in which the input data is represented.

We propose the following bi-level optimization procedure:

\begin{enumerate}[label=Step \arabic*. , wide=0pt, font=\bfseries]
\item Optimizing all encoder and decoder weights, we train the combined neural network architecture to identify the `common' subspace of the data by minimizing
 \begin{equation}\label{eqn:objective_1}
     \mcL_1=\mcL_\text{reconstruction}+\mcL_\text{common}
 \end{equation}
 
\item After fixing the common encoder weights $(\mfe_1^c, \mfe_2^c)$, we add the orthogonality constraint between them and their respective uncommon maps ($\mfe_1^u,\mfe_2^u$), to now \textit{disentangle} the uncommon component from the latent data representation. That is, we minimize:
	\begin{equation}\label{eqn:objective_2}
 \mcL_2 = \mcL_\text{reconstruction}+\mcL_\text{orthogonality}
	\end{equation}
\end{enumerate}

A detailed description of the proposed algorithm is presented in \cref{app:algs}. Intentionally separating the weights of the common encoder component and uncommon encoder component for each system (as shown in \cref{fig:architecture}) facilitates the process, in our experience, by allowing us to fix the common component weights \textit{after} identifying the common subspace in step 1. 

While, in principle, the proposed objectives do not compete, and can all three be minimized simultaneously, we have empirically observed that process to be less stable/efficient. Furthermore, minimizing $\mcL_\text{orthogonality}$ is only sensible \textit{after} the common variables have been (at least approximately) identified; otherwise it is possible for this objective to initially steer the optimization to a `bad' local minimum.

Finally, we note that minimizing the inner products of $\mcL_\text{orthogonality}$ is an attempt to \textit{block-diagonalize the metric tensor} of the systems observed by each individual sensor, in the space where the observation occurs. The common and uncommon systems are represented by their own blocks, and we penalize the $\ell^2$ norm of the off-block-diagonal entries, at each point where we have an observation. This form of disentanglement \textbf{is differential and local}, in contrast to \textbf{statistical versions} which measure conditional dependence between variables (e.g. nonlinear Independent Component Analysis \cite{hyvarinen1999ICA} or $\beta$-Variational Autoencoders \cite{higgins2017betavae}). Fundamentally what we achieve (for each sensor) is a latent parametrization which is a cross-product of two submanifolds. The effect of the orthogonality constraint is, we believe, clearly illustrated in \cref{fig:ex_LC_results}.

We demonstrate the capabilities of the proposed architecture and algorithm in the numerical examples of the following sections.

\subsection{Probabilistic Learning on Manifolds}
\label{sec:plom}

Probabilistic Learning on Manifolds and its variants (PLoM)~\cite{soize2016data, soize2022probabilistic, soize2020physics, giovanis2026enabling} generates samples whose distribution concentrates on a low-dimensional manifold embedded in a high-dimensional space. Given the dataset matrix $[\hat{u}_{d_u}] \in \mathbb{M}_{d_u,N}$ (with $N$ samples in $\mathbb{R}^{d_u}$), the goal is to generate statistically consistent replicas that preserve both the empirical distribution and the manifold geometry. After normalization, the data are decomposed by principal component analysis (PCA) via
\[
[\mathbf{c}] \boldsymbol{\phi}_k = \mu_k \boldsymbol{\phi}_k,\qquad 
\boldsymbol{\phi}_k^\top \boldsymbol{\phi}_k=1,
\]
where $[\mathbf{c}]$ is the correlation matrix. Retaining the leading $\nu$ modes in $[\boldsymbol{\Phi}]$, we write the reduced representation
\begin{equation}\label{eq:pca}
[\hat{\mathbf{U}}]=[\underline{\hat{u}}]+[\boldsymbol{\Phi}][\mu]^{1/2}[\mathbf{H}],
\end{equation}
with reduced coordinates $[\eta_d]=[\mu]^{-1/2}[\boldsymbol{\Phi}]^\top\big([\hat{u}_{d_u}]-[\underline{\hat{u}}]\big)$. A nonparametric pdf for $\mathbf{H}$ is learned by KDE using an isotropic Gaussian kernel:
\begin{equation}\label{eq:pdf}
p_{\mathbf{H}}(\boldsymbol{\eta})=\frac{1}{N}\sum_{j=1}^N 
\frac{1}{(2\pi \hat{s}_\nu^2)^{\nu/2}}
\exp\!\left(-\frac{\|\boldsymbol{\eta}^{d,j}-\boldsymbol{\eta}\|^2}{2\hat{s}_\nu^2}\right),
\end{equation}
with data-adaptive bandwidths
\[
s_\nu=\left\{\frac{4}{N(2+\nu)}\right\}^{1/(\nu+4)},\qquad 
\hat{s}_\nu=\frac{s_\nu}{\sqrt{s_\nu^2+\frac{N-1}{N}}}.
\]
Assuming independence across samples yields $p_{[\mathbf{H}]}([\eta])=\prod_{j=1}^N p_{\mathbf{H}}(\boldsymbol{\eta}^j)$. To identify the intrinsic geometry, Diffusion Maps~\cite{coifman2006diffusion} are applied to $\{\boldsymbol{\eta}^{d,j}\}_{j=1}^N$ using the kernel
\[
k_\varepsilon(\boldsymbol{\eta},\boldsymbol{\eta}')=
\exp\!\left(-\frac{\|\boldsymbol{\eta}-\boldsymbol{\eta}'\|^2}{4\varepsilon}\right).
\]
After standard normalization, we solve
\begin{equation}
[\mathbb{P}_S]\boldsymbol{\varphi}_\alpha=\lambda_\alpha\boldsymbol{\varphi}_\alpha,
\end{equation}
and define diffusion coordinates (at diffusion time $\kappa$) as
\begin{equation}
\mathbf{g}_\alpha=\lambda_\alpha^\kappa [b]^{-1/2}\boldsymbol{\varphi}_\alpha.
\end{equation}
Sampling on the learned manifold is performed via an It\^o stochastic differential equation (ISDE) 
\begin{align}
d[\mathbf{U}(r)]&=[\mathbf{V}(r)]\,dr,\\
d[\mathbf{V}(r)]&=[\mathcal{L}([\mathbf{U}(r)])]\,dr-\tfrac12 f_0[\mathbf{V}(r)]\,dr+\sqrt{f_0}\,d[\mathbf{W}(r)],
\end{align}
and reduced with a Galerkin projection onto the diffusion basis:
\begin{align}\label{eq:red_ito}
d[\mathcal{Z}(r)]&=[\mathcal{Y}(r)]\,dr,\\
d[\mathcal{Y}(r)]&=[\mathcal{L}([\mathcal{Z}(r)])]\,dr-\tfrac12 f_0[\mathcal{Y}(r)]\,dr+\sqrt{f_0}\,d[\mathcal{W}(r)].
\end{align}
With initial conditions $[\mathcal{Z}(0)]=[\eta_d][a]$ and $[\mathcal{Y}(0)]=[\mathcal{N}][a]$, the reduced trajectory is integrated (St\"ormer--Verlet). Generated samples are then reconstructed in the ambient space by
\begin{equation}
[\eta_s]=[\mathcal{Z}(l,\rho)][g]^\top,\qquad 
[\hat{u}_s^l]=[\underline{\hat{u}}]+[\boldsymbol{\Phi}][\mu]^{1/2}[\eta_s^l].
\end{equation}
In this work, we use the original PLoM formulation as discussed in \cite{soize2016data}.

\subsection{Score-based Diffusion Model}
\label{sec:score_diffusion}

In contrast to the PLoM framework—which constructs a reduced-order probabilistic model concentrated on a learned low-dimensional manifold—the second approach leverages score-based diffusion models \cite{song2021maximum}, with a specific focus on the formulation presented in~\cite{liu2024diffusion, GIOVANIS2025118266}. Here, we consider the dataset of i.i.d. samples $\{\hat{\vb{u}}^1, \dots, \hat{\vb{u}}^N\} \subset \mathbb{R}^{d_u}$ drawn from an unknown probability density $p_{\hat{\vb{u}}}$. The goal is to learn a generative model that enables direct sampling from this complex high-dimensional distribution without explicitly reducing its dimensionality. We define a continuous-time diffusion process $\{\vb{u}(t)\}_{t=0}^1$, where $\vb{u}(0) \sim p_{\hat{\vb{u}}}$ and $\vb{u}(1) \sim \mathcal{N}(\vb{0}, \vb{I}_{d_u})$. The forward SDE is defined as: 

\begin{equation}\label{eq:forwardSDE} \text{d}\vb{u}_t = g(\vb{u}_t, t) \, \text{d}t + \sigma(t) \, \text{d}W_t, 
\end{equation} 
with drift $g(\vb{u}, t) = b(t)\vb{u}_t$ and diffusion $\sigma^2(t) = \text{d}\beta^2_t/\text{d}t - 2 (\text{d}\log \alpha_t/\text{d}t) \beta^2_t$. We use $\alpha_t = 1 - t$, $\beta_t^2 = t$ to guarantee $\vb{u}(1) \sim \mathcal{N}(\vb{0}, \vb{I})$. Given this parameterization, the conditional density satisfies: 

\begin{equation} p(\vb{u}_t | \vb{u}_0) = \mathcal{N}(\alpha_t \vb{u}_0, \beta_t^2 \vb{I}). 
\end{equation} 

The reverse SDE is $\text{d}\vb{u}_t = \left(g(\vb{u}_t, t) - \sigma^2(t) \, \nabla_{\vb{u}_t} \log p_t(\vb{u}_t)\right) \text{d}t + \sigma(t) \, \text{d}\overleftarrow{W}_t,$ which requires the score function $S(\vb{u}_t, t) = \nabla_{\vb{u}_t} \log p_t(\vb{u}_t)$. Using the law of total probability, we write: \begin{equation} S(\vb{u}_t, t) = \sum_{n=1}^{N_m} \frac{\vb{u}_t - \alpha_t \hat{\vb{u}}^n}{\beta_t^2} \, \overline{w}_t(\vb{u}_t, \hat{\vb{u}}^n), \end{equation}

with weights approximated via mini-batch MCS \cite{liu2024diffusion}: \begin{equation} \overline{w}_t(\vb{u}_t, \hat{\vb{u}}^n) = \frac{p(\vb{u}_t | \hat{\vb{u}}^n)}{\sum_{m=1}^{N_m} p(\vb{u}_t | \hat{\vb{u}}^m)}. \end{equation} To reduce sampling complexity, we approximate the reverse-time SDE with an ordinary differential equation (ODE) governed by the Liouville equation: \begin{equation} \text{d}\vb{u}_t = \left(g(\vb{u}_t, t) - \frac{1}{2}\sigma^2(t) S(\vb{u}_t, t)\right) \text{d}t. \end{equation} This smoother flow enables training a neural network that maps $\vb{y}_m \sim \mathcal{N}(\vb{0}, \vb{I})$ to labeled data $\hat{\vb{u}}_m$ using the training set $D_{\text{train}} = \{(\vb{y}_m, \hat{\vb{u}}_m)\}_{m=1}^M$. Once trained, sampling is direct without solving the SDE or the ODE. 

\section{Computational Examples}\label{sec:examples}

We first study a synthetic example arising in a dynamical systems context, before revisiting the `bobble-head problem' introduced in \cite{lederman2014}. The latter is a dynamical system in disguise but, nevertheless, showcases the ability of our architecture to identify common/uncommon underlying structures, that can be further used to generate high-dimensional images. Finally, we discuss an example where the two sets of observations are separated by a consistent time shift. This allows us to effectively learn a time-stepping map (a ``world model") for an evolving dynamical system.

For additional details regarding the data generation, see \cref{app:net_data}.

\begin{exmp}\label{exmp:Torus}
\textbf{(Torus)} The data sets for this first computational example consist of samples of oscillator trajectories. These are combined using a random matrix, so that each observation contains information from both the common and uncommon systems. Each $\mcU,\mcV,\mcC$ are closed orbits (limit cycle-type trajectories). Both common and uncommon latent spaces are two-dimensional. Note that this is a minimal realization of the rotating bobblehead example in \cite{lederman2014}.
 In \cref{fig:ex_LC_results} we visualize the resulting embedding of the data after the first (top row) and after the second (bottom row) levels of our optimization are completed. In the top row we see that the common subspaces (blue) are `correctly' identified to be topological circles, while the uncommon ones (red and green) are random projections of the overall higher-dimensional (toroidal) data. After imposing the orthogonality constraints between the common and uncommon subspaces for each system, we recover the `decoupled' circles in both cases. Each image is colored by the corresponding intrinsic variable $\theta_i$ (corresponding to the `correct' coordinates on each subspace, up to diffeomorphism); these known \textit{a priori} since the data set was generated synthetically, and should therefore vary smoothly upon proper convergence of the algorithm.

    \begin{figure}[h]
		\begin{subfigure}[b]{0.24\textwidth}
			\includegraphics[width=\textwidth]{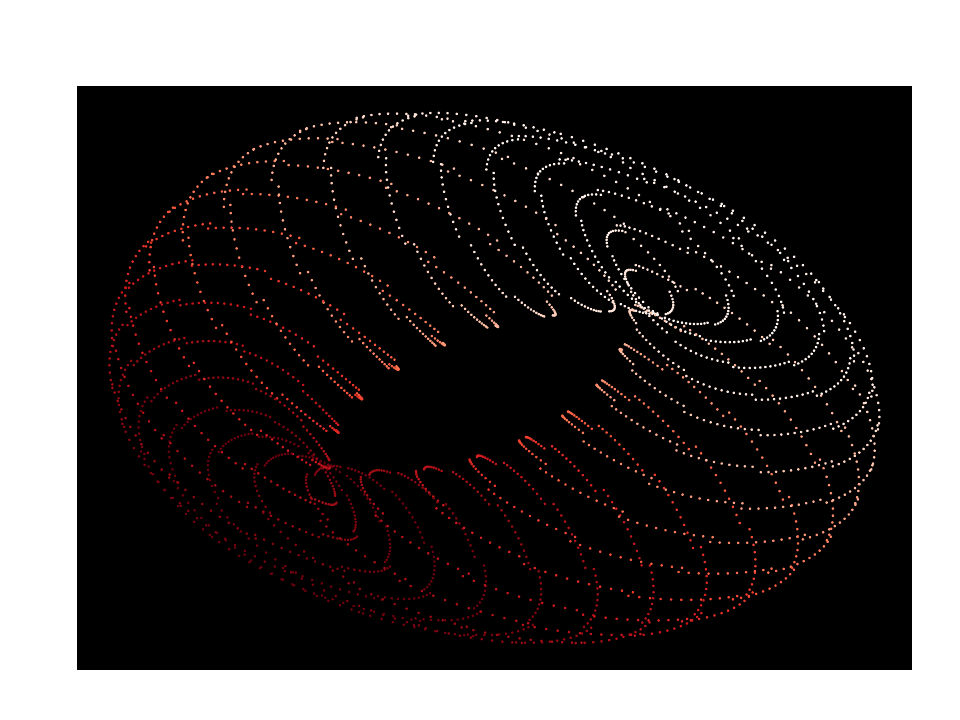}
			\caption*{Uncommon $S_1$}
		\end{subfigure}
		\begin{subfigure}[b]{0.24\textwidth}
			\includegraphics[width=\textwidth]{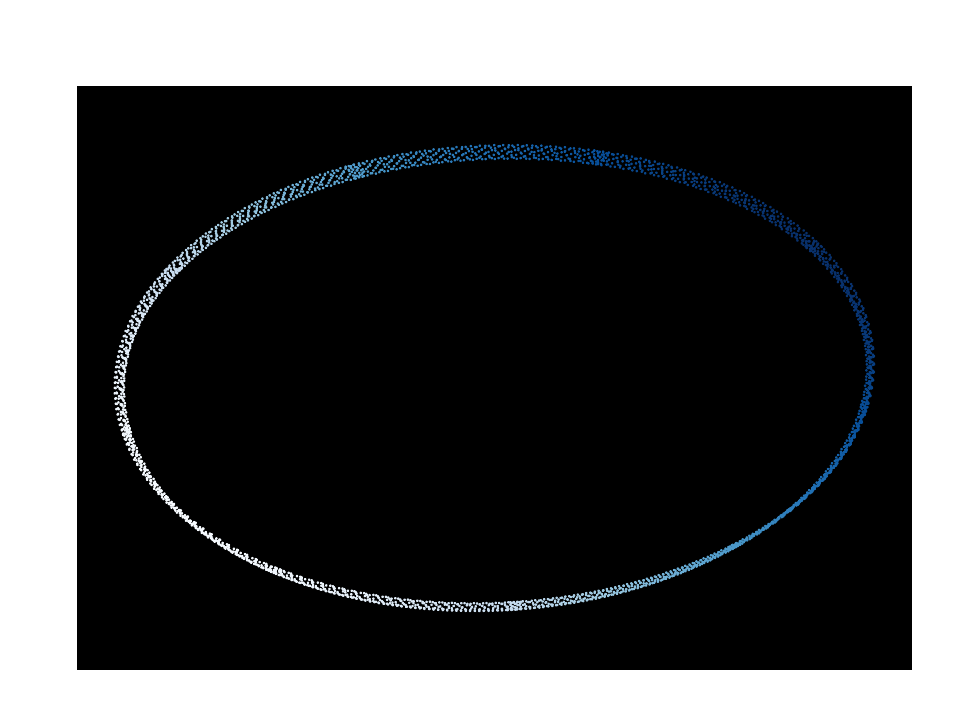}
			\caption*{Common $S_1$}
		\end{subfigure}
		\begin{subfigure}[b]{0.24\textwidth}\includegraphics[width=\textwidth]{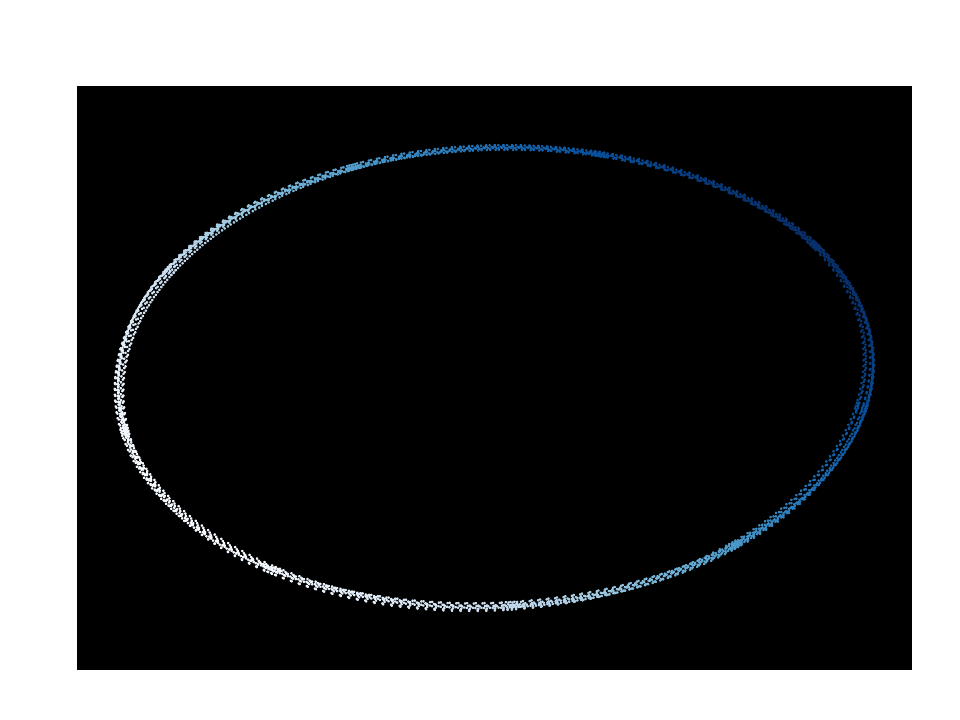}
			\caption*{Common $S_2$}
		\end{subfigure}
		\begin{subfigure}[b]{0.24\textwidth}
			\includegraphics[width=\textwidth]{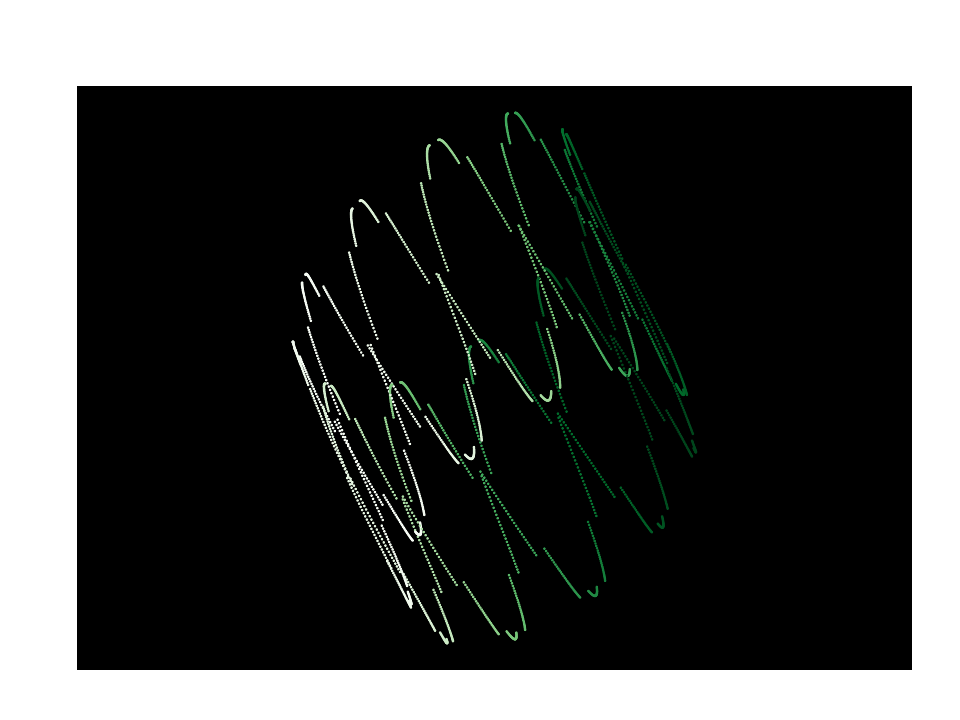}
			\caption*{Uncommon $S_1$}
		\end{subfigure}\\
		\begin{subfigure}[b]{0.24\textwidth}
			\includegraphics[width=\textwidth]{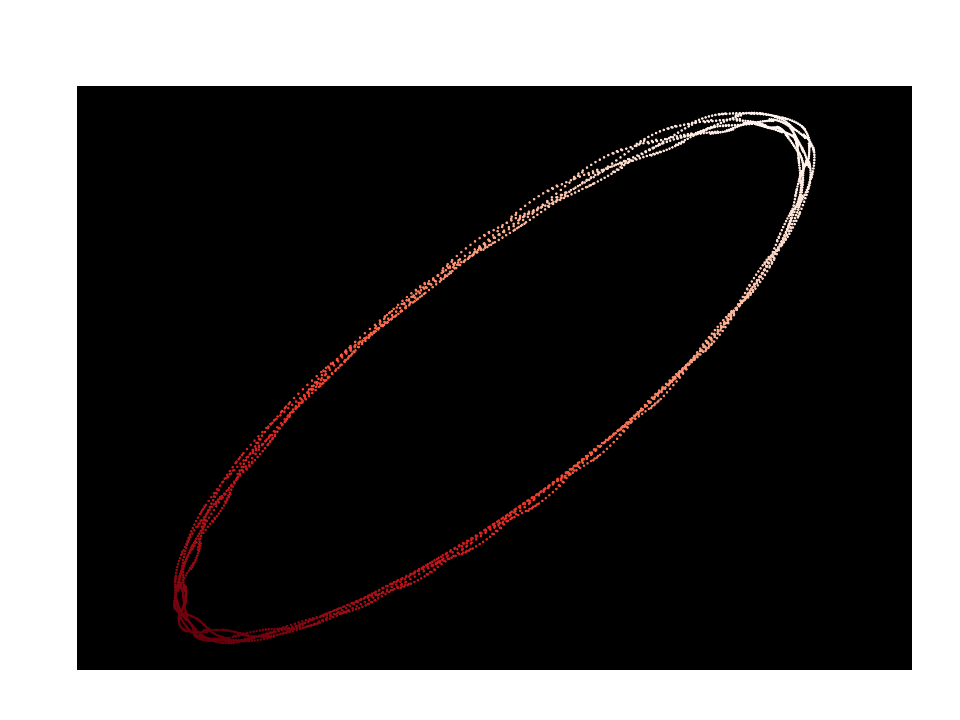}
		\end{subfigure}
		\begin{subfigure}[b]{0.24\textwidth}
			\includegraphics[width=\textwidth]{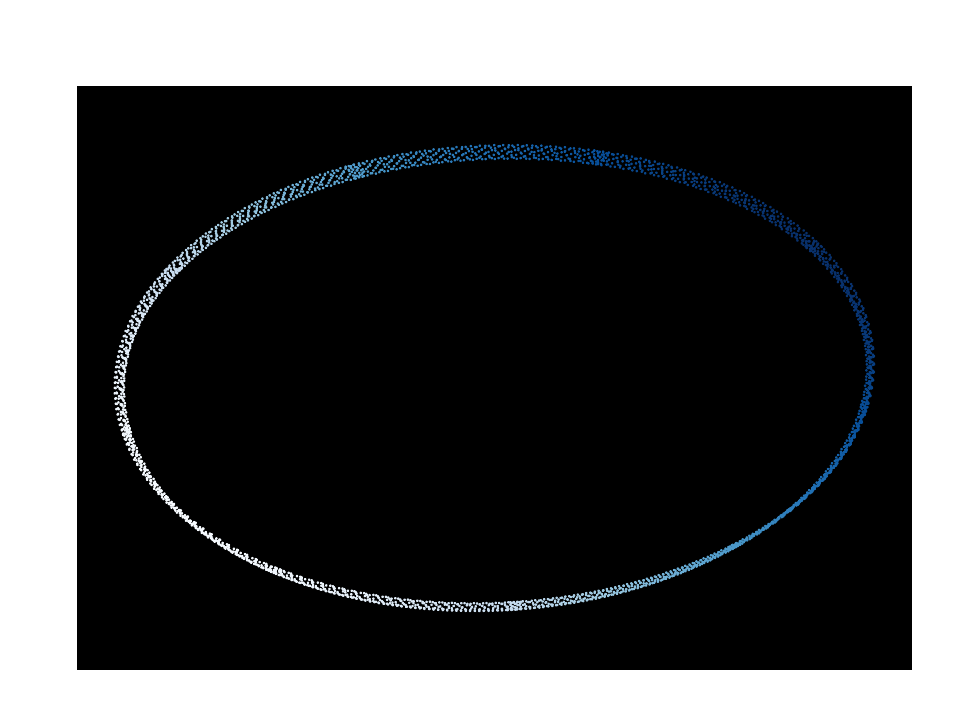}
		\end{subfigure}
		\begin{subfigure}[b]{0.24\textwidth}\includegraphics[width=\textwidth]{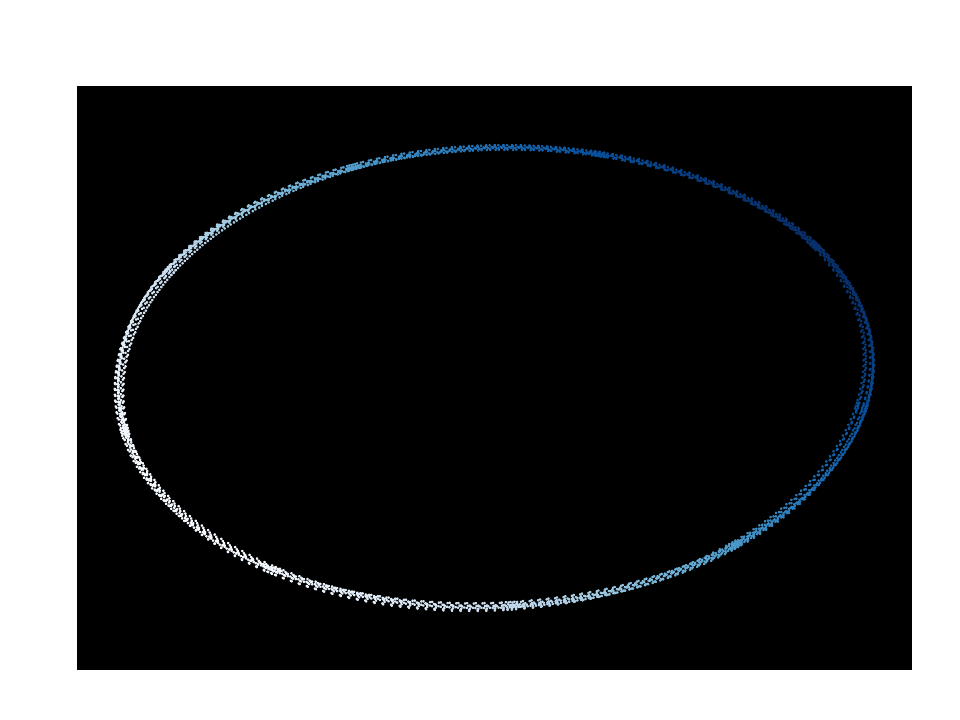}
		\end{subfigure}
		\begin{subfigure}[b]{0.24\textwidth}
			\includegraphics[width=\textwidth]{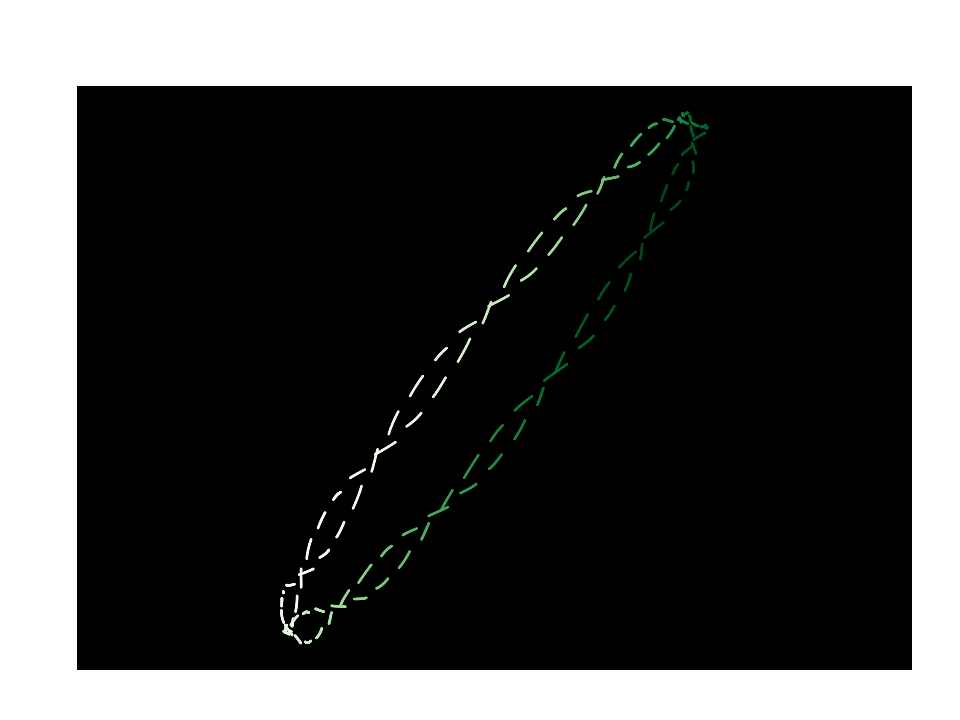}
		\end{subfigure}
		\caption{Latent embeddings produced for a single run of the optimization algorithm applied to \cref{exmp:Torus}. (\textbf{Top Row}) Result after successfully terminating the first level optimization (no orthogonality between the common and uncommon components). (\textbf{Bottom Row}) Resulting embedding after the second level of optimization (with orthogonality imposed).}
        \label{fig:ex_LC_results}
	\end{figure}
	To produce the corresponding figures, we use 54\% of the data set for training, 36\% for validation and 10\% for testing. We terminate the first optimization step with training and validation losses of $\num{1.1e-5},\num{1.0e-5}$ respectively, and the second optimization step with losses of $\num{5.0e-4},\num{9.1e-5}$ respectively. The final test loss reported was $\num{1.0e-4}$.
	
\end{exmp}

\begin{figure}[h]
\centering
\begin{subfigure}[b]{0.30\textwidth}
    \centering
    \includegraphics[width=\textwidth]{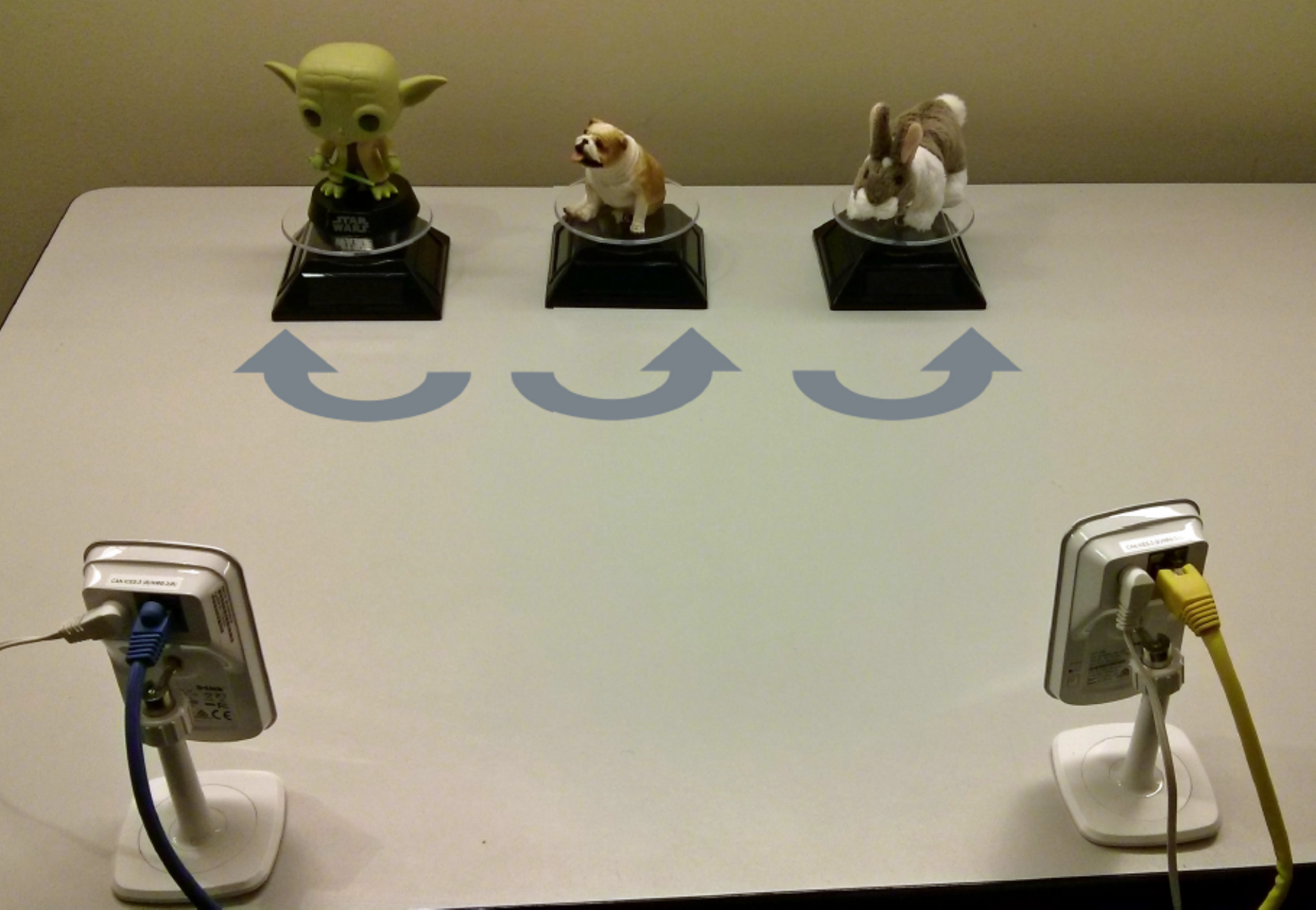}
    \caption{}
\end{subfigure}
\quad
\begin{subfigure}[b]{0.58\textwidth}
    \centering
    \includegraphics[width=\textwidth]{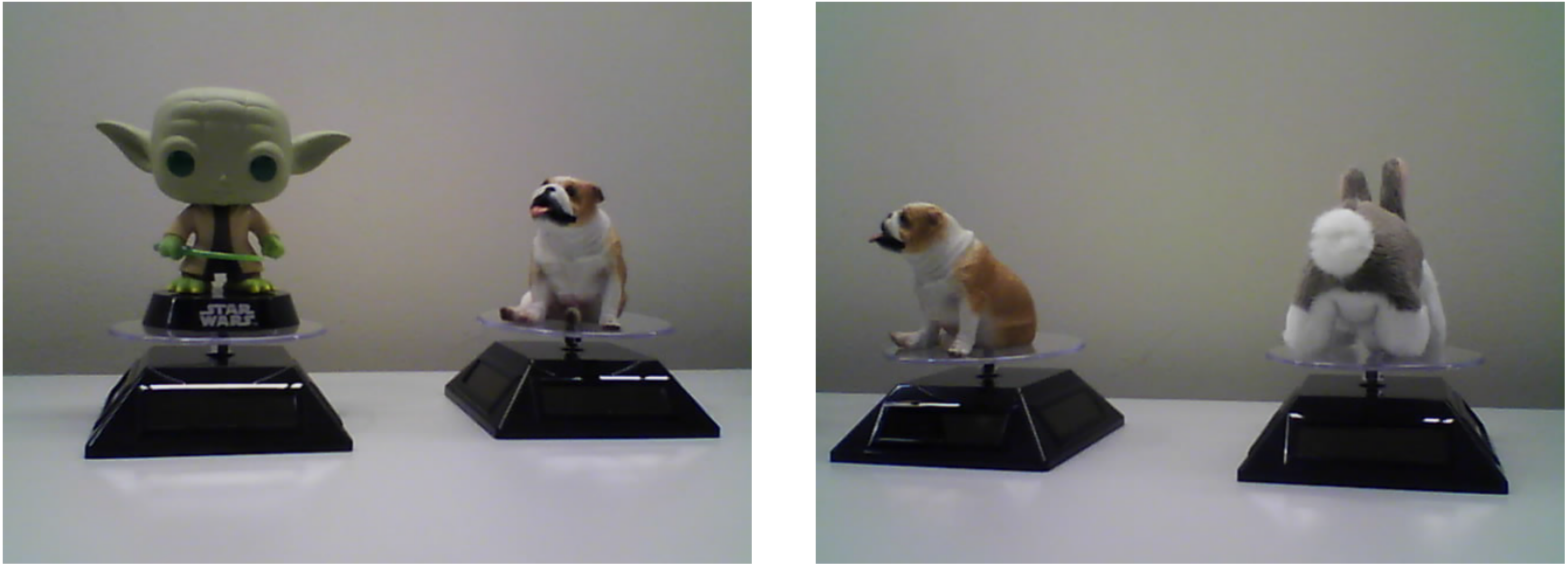}
    \caption{}
\end{subfigure}
\caption{Bobblehead simulation setup presented in \cite{lederman2018}. (We use the figures available in \cite{sroczynski2024learninglearnheterogeneousobservations}). (a) gives a `bird's-eye' view of the setup, with the two cameras observing the bobblehead system; each bobblehead rotates independently with an unknown frequency. (b) shows each camera's point of view at a specific point in time. The bulldog is `common' between them but observed at different angles, while Yoda and the rabbit are `uncommon'.}
\label{fig:bobblehead_setup}
\end{figure}

\begin{exmp}[Bobbleheads]\label{exmp:bobbleheads}
    Lastly, we consider the prototypical situation described in \cref{exmp:intro} where the roles of Alice, Bob, and Carol are played by rotating bobbleheads (\cref{fig:bobblehead_setup} Yoda $(\mcU)$, a Bulldog $(\mcC)$, and a Rabbit $(\mcV)$). These appeared in the original Alternating Diffusion work of \cite{lederman2018}. We note that geometrically, the setting is identical to that of Example 1, with the difference being that we have to work with (higher-dimensional) images instead of the (lower embedding-dimensional) closed-curve trajectories considered previously.

    Our training set consists of 2500 pairs of RGB images in $320\times250$ pixels each, which are converted to grayscale and compressed using their first 60 principal components. The PCA projection gives us a considerably smaller embedding dimension to work in computationally, but also `entangles' the coordinates (pixels) of the three systems, with each PCA coordinate being a function of the pixels corresponding to \textit{both} bobble-heads each sensor observes. One may in principle use random projections or a wavelet expansion instead and achieve a qualitatively similar result.

    We start by training a variant of the proposed architecture \cref{sec:alt_architecture}, which we find to be more robust in identifying the common variable subspace. After identifying the common subspace, we train the original architecture (simply by switching the decoder inputs to what is depicted in \cref{fig:architecture}) to also identify the uncommon systems. The final representations are portrayed in \cref{fig:Image_embeddings}.

    \begin{figure}[ht!]
            \begin{subfigure}[b]{0.24\textwidth}
			\includegraphics[width=\textwidth]{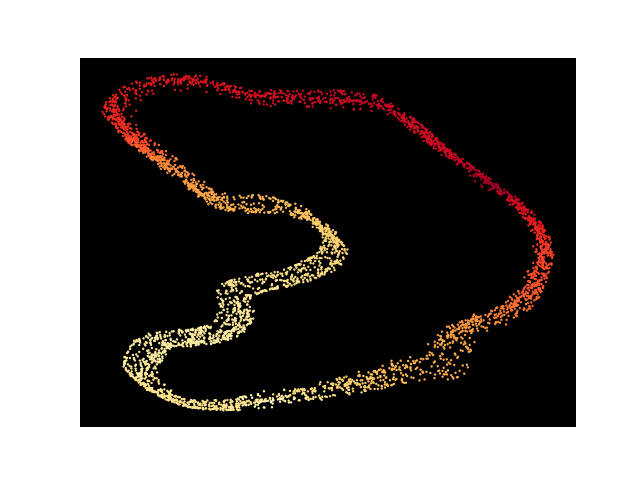}
			\caption*{Uncommon $S_1$}
		\end{subfigure}
            \begin{subfigure}[b]{0.24\textwidth}
			\includegraphics[width=\textwidth]{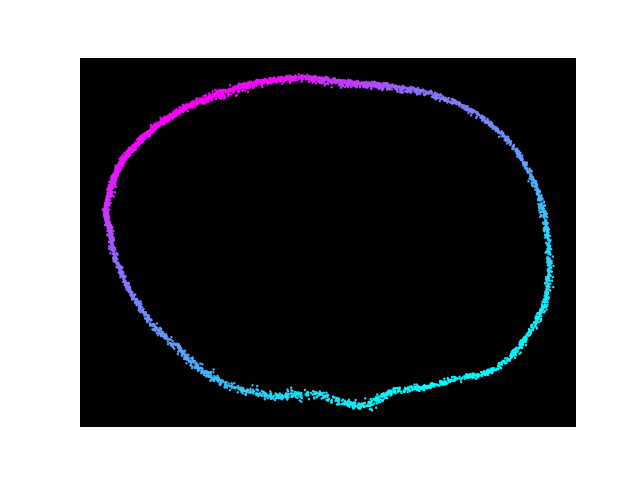}
			\caption*{Common $S_1$}
		\end{subfigure}
		\begin{subfigure}[b]{0.24\textwidth}
                \includegraphics[width=\textwidth]{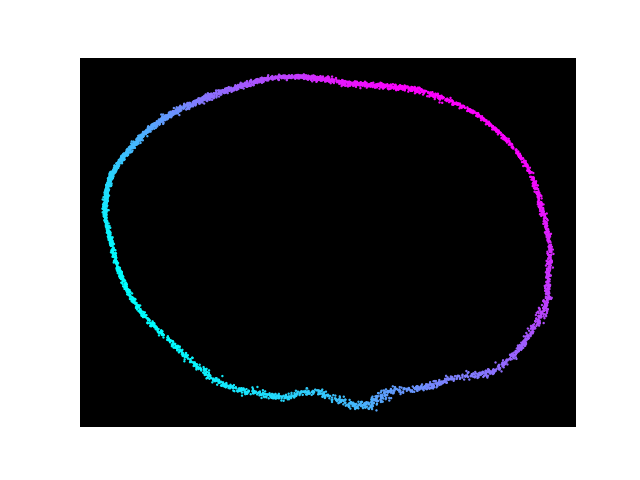}
			\caption*{Common $S_2$}
		\end{subfigure}
		  \begin{subfigure}[b]{0.24\textwidth}
			\includegraphics[width=\textwidth]{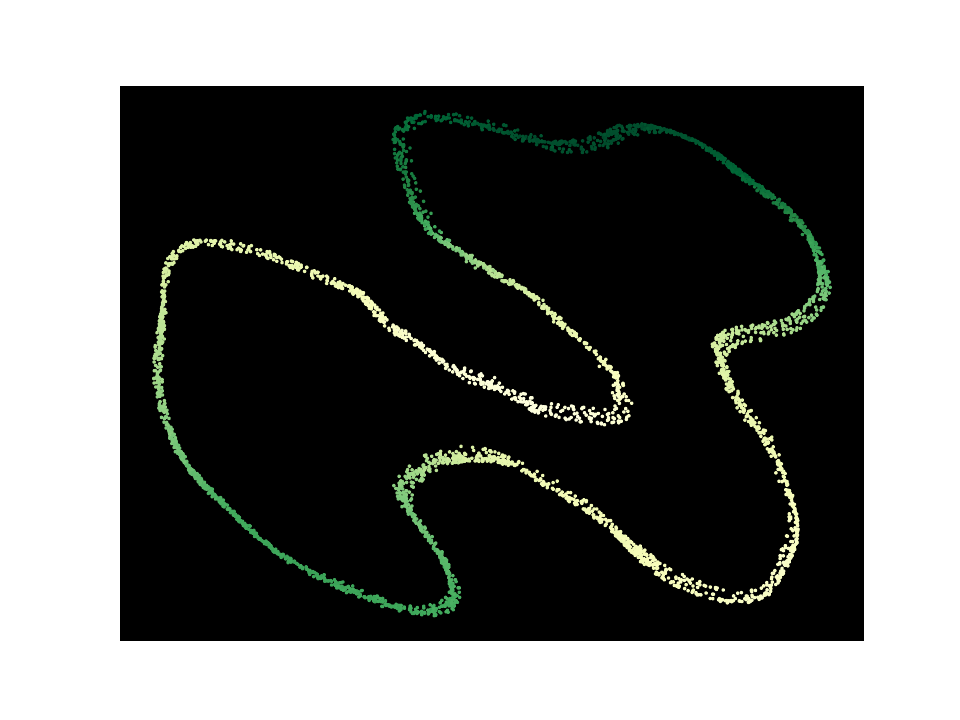}
			\caption*{Uncommon $S_2$}
		\end{subfigure}
		\caption{Latent embeddings produced by a single run of the optimization algorithm applied to \cref{exmp:bobbleheads}. Each individual system corresponds to a limit cycle, with points now representing images instead of oscillator trajectories. Here the ground-truth angles are not known; instead , the `common' systems (middle-left, middle-right) are respectively colored by the two common coordinates approximated using the Alt-DMap algorithm of \cite{lederman2014,lederman2018}. The Uncommon systems are colored based on one of the two predictions of the Output-Informed Alt-DMap algorithm of \cite{sroczynski2024learninglearnheterogeneousobservations}.}
		\label{fig:Image_embeddings}
	\end{figure}

    \subsection*{Level Sets}
    Given an observation of camera 1, we know what the common features (the bulldog) in a simultaneous camera 2 observation will look like - but we do not know what the uncommon component of that observation will be. In fact, there exists an entire level set of camera 2 observations consistent with a particular camera 1 observation. 
    
     We further demonstrate the usefulness of our embeddings in producing such level sets. This is accomplished by generating `artificial' images, exploiting the trained AE architecture (\cref{fig:Image_generated}). We first embed an `original' image in our computed, disentangled latent space, where the common and uncommon coordinates are represented by white `X's. Then, after fixing either its \textbf{(a)} common coordinate or \textbf{(b)} uncommon coordinate, we randomly sample separate points in the alternate coordinate and feed the combination through the decoder. The reconstructed images are plausible observations, consistent with a fixed orientation for the common (resp. uncommon) bobble-head. Such a level set construction can be thought of as an extension of the Gappy-POD \cite{koronaki2023POD, papavasileiou2023POD}, Gappy-Diffusion Maps \cite{martin2023physics} and Gappy-LOCA \cite{peterfreund2023gappyloca} computational workflows.

     \begin{figure}[H]
         \centering
         \includegraphics[trim={0 0 0 2cm}, width=\textwidth]{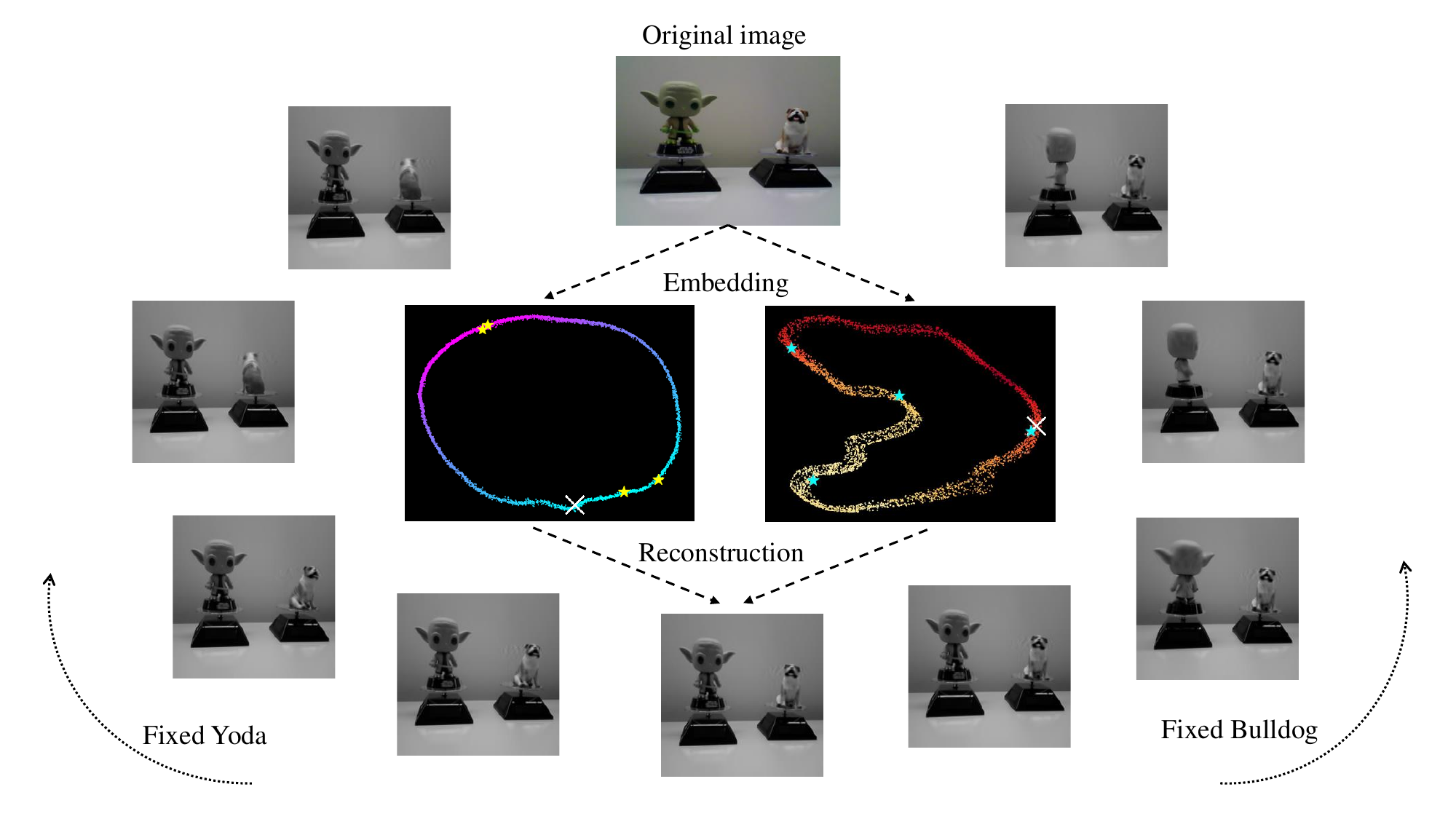}
         \caption{Images generated using the converged AE network. The larger white `X' denote the embedding of the original image, which is reconstructed using the decoder approximation of its first 60 principal components. Then, we generate images consistent with either a fixed Bulldog (fixed common system, counter-clockwise) or even with a fixed Yoda (fixed uncommon system, clockwise) by fixing the corresponding common or uncommon coordinate and randomly varying over choices of points in the other coordinate (yellow $\star$ for a varying bulldog and cyan $\star$ for a varying yoda).}
         \label{fig:Image_generated}
     \end{figure}
\end{exmp}

\subsection*{`Causal' Learning}\label{sec:causal}
 Throughout the examples of \cref{sec:examples}, we have assumed that each of the two sensors takes pictures (samples) \textit{simultaneously}; this is natural in settings similar to \cref{exmp:intro}.

 However, it is possible to have a constant time difference $\Delta t$, or lag, between the observations of the two sensors, and \textit{still} be able to achieve the objectives of \cref{sec:Problem_Statement}. That is, a diffeomorphism $\varphi$ as in \cref{eqn:diffeo} may still exist. If, for each observation of Sensor 1 at time $t$, sensor 2 makes an observation at time $t+\Delta t$, and the dynamical system evolves smoothly and deterministically, the state (of the common system $\mathcal{C}$) at time $t+\Delta t$ is fully determined by integrating the state of the system at time $t$. This is the \textit{flow map}:
 \begin{equation*}
    \Phi^\mcC_{\Delta t}:(c_{1},...,c_{d_c})(t)\mapsto(c_{1},...,c_{d_c})(t+\Delta t).
 \end{equation*}
 Of course, as $\Delta t$ increases, numerical sensitivity issues arise when chaotic dynamics are involved.

To demonstrate the ability of the architecture to handle time-delayed data, we construct a simple example: We let $\mcC$ be a common limit cycle, and $\mcU$ be a R\"{o}ssler attractor. The first sensor \textit{only} observes the limit cycle at time $t$, while the second sensor observes the limit cycle at time $t+\Delta t$, as well as the uncommon system (here, $\Delta t=200$ time units). 

In this case, we do not scramble the data observed by the second sensor. Thus, observations have the form:

\begin{gather*}
    S_1:(c_{1,t},c_{2,t})\\
    S_2:(c_{1,t+\Delta t},c_{2,t+\Delta t}, u_{1,t+\Delta t},u_{2,t+\Delta t},u_{3,t+\Delta t})
\end{gather*}

\begin{figure}
    \centering
    \includegraphics[width=0.5\linewidth]{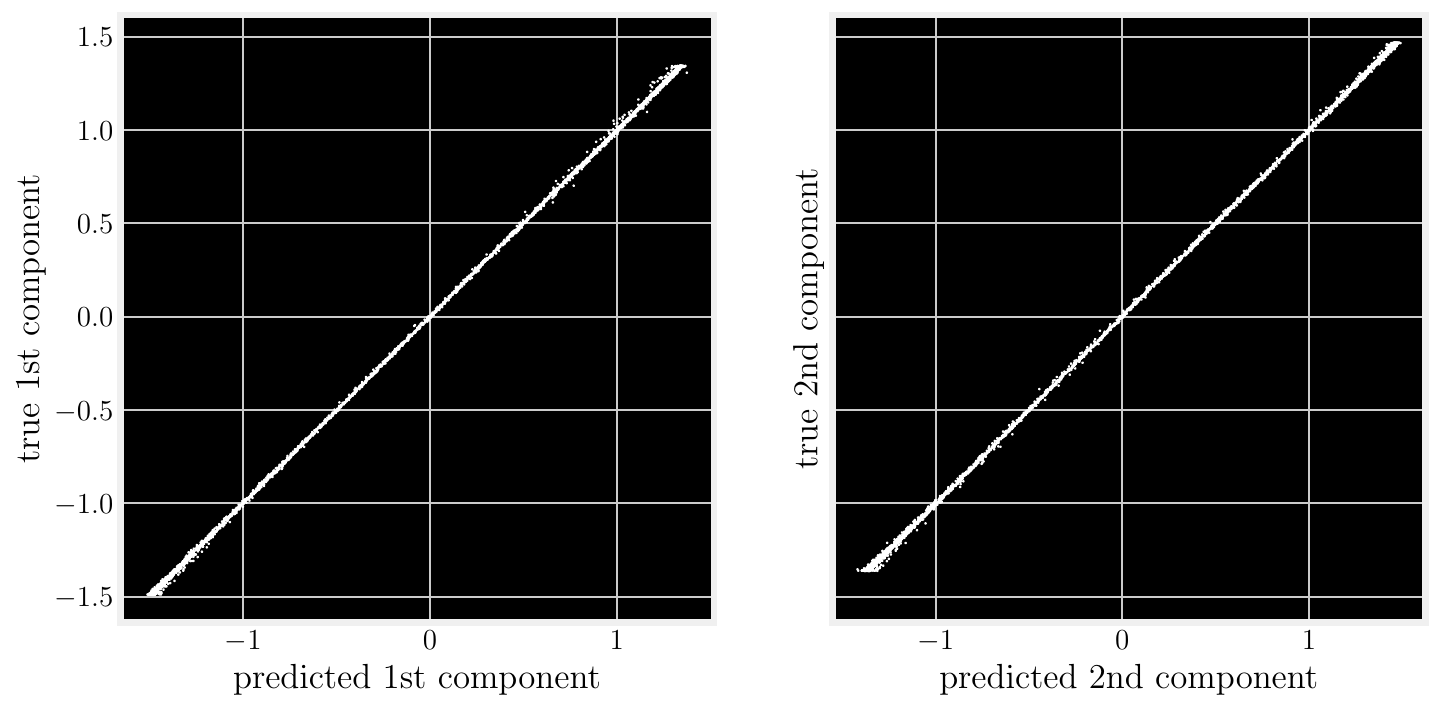}
    \caption{Relationship between the predicted ($\hat{\vb{c}}_{t}$) and true state ($\vb{c}_{t}$) of the common dynamical system at time $t$.}
    \label{fig:causal}
\end{figure}

We train the architecture as previously proposed, to learn a common latent space and to disentangle the uncommon system that the second sensor observes. Then, encoding the observation of the second sensor (at time $t+\Delta t$) and using the decoder of the first sensor, we deduce (``postdict")  the state of the common dynamical system at time $t$. Mathematically this is described as:

\begin{align}
    \hat{\vb{c}}_{t}=\mfd_1\circ\pi_{1,2}\circ\mfe_2\qty(\vb{c}_{t+\Delta t}, \vb{u}_{t+\Delta t})
\end{align}
where $\pi_{1,2}$ denotes a projection on the first two (here, the `common') coordinates. In \cref{fig:causal} we show that the predicted ($\hat{\vb{c}}_{t+\Delta t}$) and true state $\vb{c}_{t+\Delta t}$ of the system at time $t+\Delta t$ match, meaning that we can accurately integrate using the trained architecture, at least for this `simple' example.

What this demonstrates is that this architecture can learn correlations between the present and future states of the dynamical system; That is, it can help establish a type of (correlational, Granger \cite{granger1969investigating}) `causality' among different asynchronous observations.


\section{Generative Learning on Uncommon Latent Spaces} \label{sec:GM}

Let $\hat{\vb{u}} = (u_1, \dots, u_{d_u}) \in \mathbb{R}^{d_u}$ and $\hat{\vb{v}} = (v_1, \dots, v_{d_v}) \in \mathbb{R}^{d_v}$ denote two generic data points in the uncommon 1 and uncommon 2 latent spaces, respectively, and let $\text{d}\hat{\vb{u}} = \text{d}u_1 \cdots \text{d} u_{d_u}$ and $\text{d}\hat{\vb{v}} = \text{d}v_1 \cdots \text{d} v_{d_v}$ be the associated Lebesgue measures. Consider two datasets, each consisting of $N$ realizations of the random vectors: $\{\hat{\vb{u}}^1, \dots, \hat{\vb{u}}^N\} \in \mathbb{R}^{d_u \times N}$ and $\{\hat{\vb{v}}^1, \dots, \hat{\vb{v}}^N\}\in \mathbb{R}^{d_v \times N}$. Each dataset has some arbitrary probability density function (pdf) ($p_{\hat{\vb{u}}}$, $p_{\hat{\vb{v}}}$) supported on a subset $S_{d_u} \subset \mathbb{R}^{d_u}$ and $S_{d_v} \subset \mathbb{R}^{d_v}$, respectively. The objective of generative models is to sample from the  unknown probability distributions. Next, we discuss two approaches for sampling,  namely Probabilistic Learning on Manifolds (PLoM) and Score-based generative models (SGMs). Throughout this section, we present the methodology using notation for the uncommon 1 latent space ($\hat{\vb{u}}$); the same applies analogously to the uncommon 2 latent space ($\hat{\vb{v}}$).

\

\subsection*{R\"{o}ssler attractor , Lorenz attractor,  Takoudis limit cycle}\label{sec:gen_results}

\label{exmp:RL}

\begin{figure}[H]
    \centering
    \includegraphics[width=0.8\linewidth]{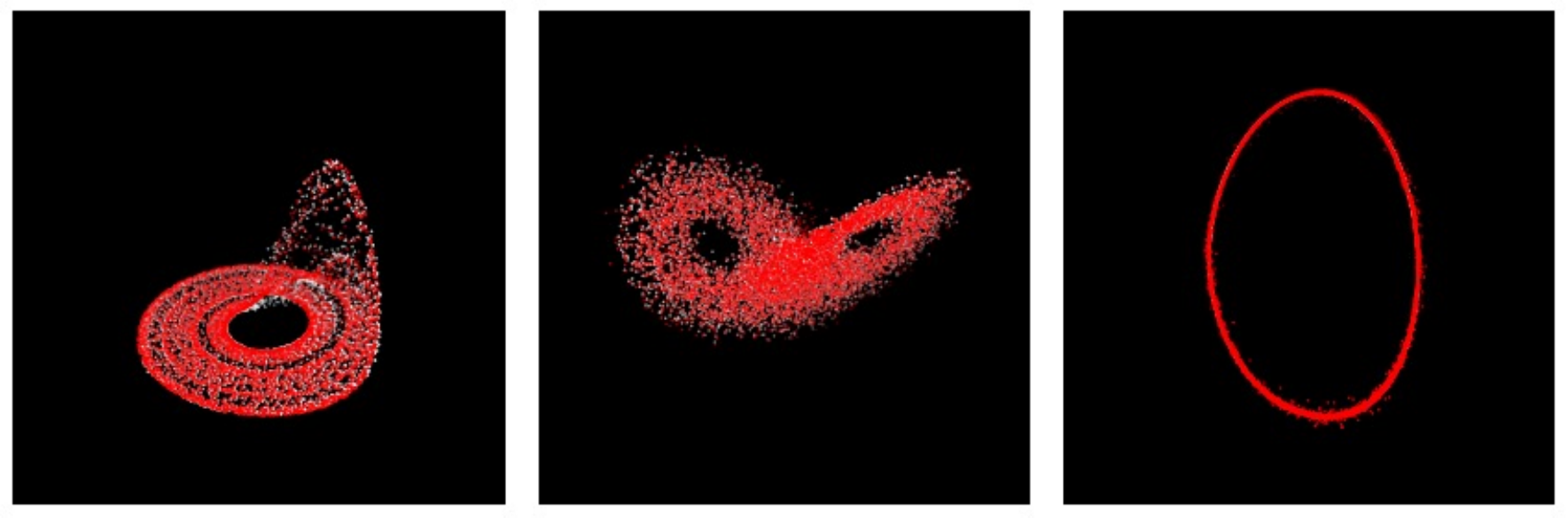}
    \caption{Points along the trajectories of the dynamical systems: The Rössler attractor (Left), the Lorenz attractor (Center) and the nonlinear limit cycle (Right)}
    \label{fig:AutoRec}
\end{figure}

To demonstrate the effectiveness of the proposed generative modeling approaches, we consider three canonical nonlinear dynamical systems: the R\"{o}ssler attractor ($\mcU$), the Lorenz attractor ($\mcV$), and the Takoudis reactor (common) limit cycle ($\mcC$)  (cf. Fig. \ref{fig:AutoRec}). These systems are well-known for exhibiting complex behavior such as chaos and multi-scale oscillations, making them ideal test cases for evaluating generative performance on structured yet high-dimensional datasets.

The data consists of samples of trajectories for the following:

\begin{itemize}
\item $\mathcal{U}$ is sampled from a R\"{o}ssler attractor:
\begin{align}
    \begin{split}
        \dot{x}&=-y-z\\
        \dot{y}&=x+\alpha y\\
        \dot{z}&=b+z(x-c)
    \end{split}
\end{align}
with $\alpha=b=0.2$ and $c=5.7$.

\item $\mathcal{V}$ is sampled from a Lorenz attractor:
\begin{align}
    \begin{split}
        \dot{x}&= \sigma(y-x)\\
        \dot{y}&=x(\rho-z)-y\\
        \dot{z}&=xy-\beta z
    \end{split}
\end{align}
with $\sigma=10$, $\beta=8/3$, $\rho=28$.

\item $\mathcal{C}$ is sampled from a nonlinear limit cycle:
\begin{align}
    \begin{split}
        \dot{x}&=\alpha_1(1-x-y)-\gamma_1x-xy(1-x-y)^2\\
        \dot{y}&=\alpha_2(1-x-y)-\gamma_2y-xy(1-x-y)^2
    \end{split}
\end{align}
with $\alpha_1=0.016$, $\gamma_1=0.001$, $\alpha_2=0.0278$, $\gamma_2=0.002$.
\end{itemize}

\begin{figure}[H]
    \centering
    \includegraphics[width=0.8\linewidth]{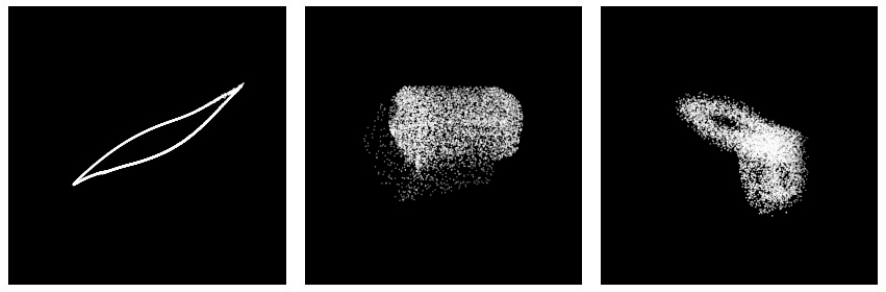}
    \caption{Latent embeddings produced by the algorithm. The common subspaces (Left) are two dimensional while the uncommon are three-dimensional for each system: Rössler (Center) and Lorenz (Right).}
    \label{fig:embedding}
\end{figure}

We sample a total of 3,000 temporally equidistant points for the data set, using accurate numerical simulation. We ``scramble'' the data using an invertible linear transformation, such that individual coordinates do not correspond to a single particular system for each sensor. These are combined using a random matrix, so that each observation contains information from both the common and uncommon systems (what we call ``dirty" observations. Thus, the common subspace is two-dimensional while the uncommon ones are three-dimensional each respectively. 
The recovered embeddings shown in Fig. \ref{fig:embedding} demonstrate the expected topological and geometric characteristics.

\subsection{Synthetic embeddings using Score-based Diffusion Model}
\label{sec:SGM_results}

We assess the generative performance of the score-based diffusion model (SGM) using 3D latent embeddings of the Rössler system. The model is trained on extracted latent representations and used to generate new samples by drawing from the learned score function. To evaluate the similarity between real and synthetic embeddings, we employ two complementary metrics: multivariate Kullback–Leibler (KL) divergence~\cite{kullback1951information} and the mean Wasserstein distance~\cite{villani2008optimal}. The former captures differences in joint probability density, while the latter quantifies geometric deviation in sample distribution, together offering a comprehensive view of model fidelity.

\begin{figure}[H]
    \centering
    \includegraphics[width=.9\linewidth]{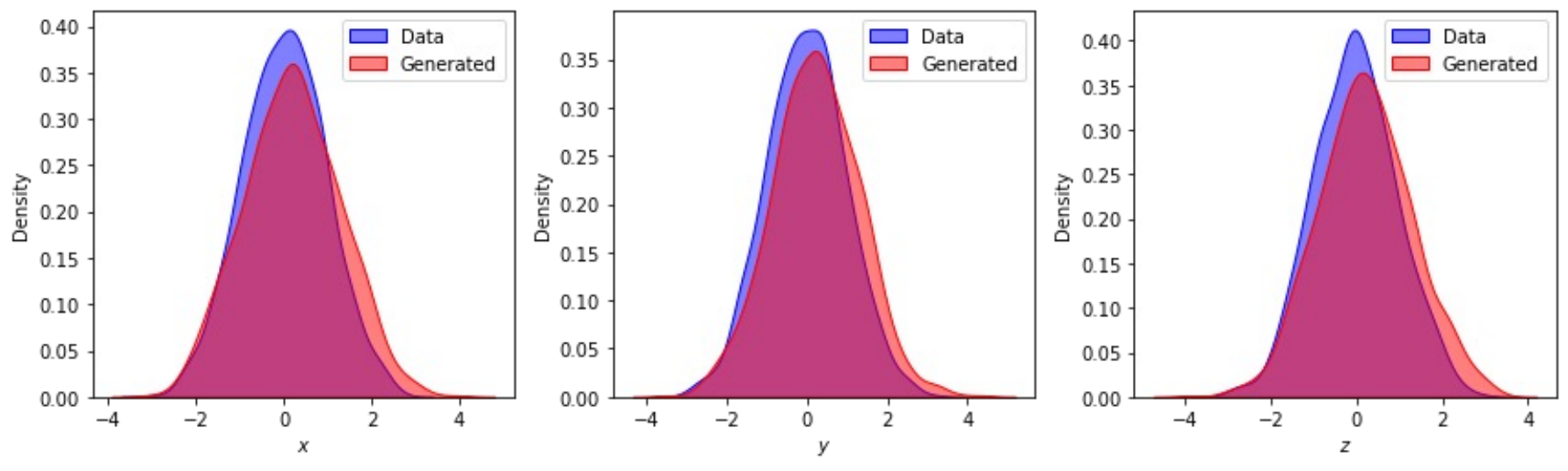}
    \caption{Rössler embeddings: Kernel density estimates of marginal distributions across latent dimensions for reference (blue) and generated (red) samples. A KL divergence of 0.3961 and Wasserstein distance of 0.2314 indicate moderate distributional divergence.}
    \label{fig:SGM_dist_Rossler}
\end{figure}

We also visualize the marginal distributions across each latent dimension using kernel density estimation (KDE), which provides an intuitive comparison of the generated and reference data (Fig.~\ref{fig:SGM_dist_Rossler}). The KL divergence between the true and synthetic Rössler embeddings is 0.3961, indicating a moderate mismatch in probabilistic structure. The mean Wasserstein distance of 0.2314 suggests that while the spatial distribution is generally preserved, subtle shifts remain, consistent with visual differences in the KDE curves.

\begin{figure}[h!]
    \centering
    \includegraphics[width=0.7\linewidth]{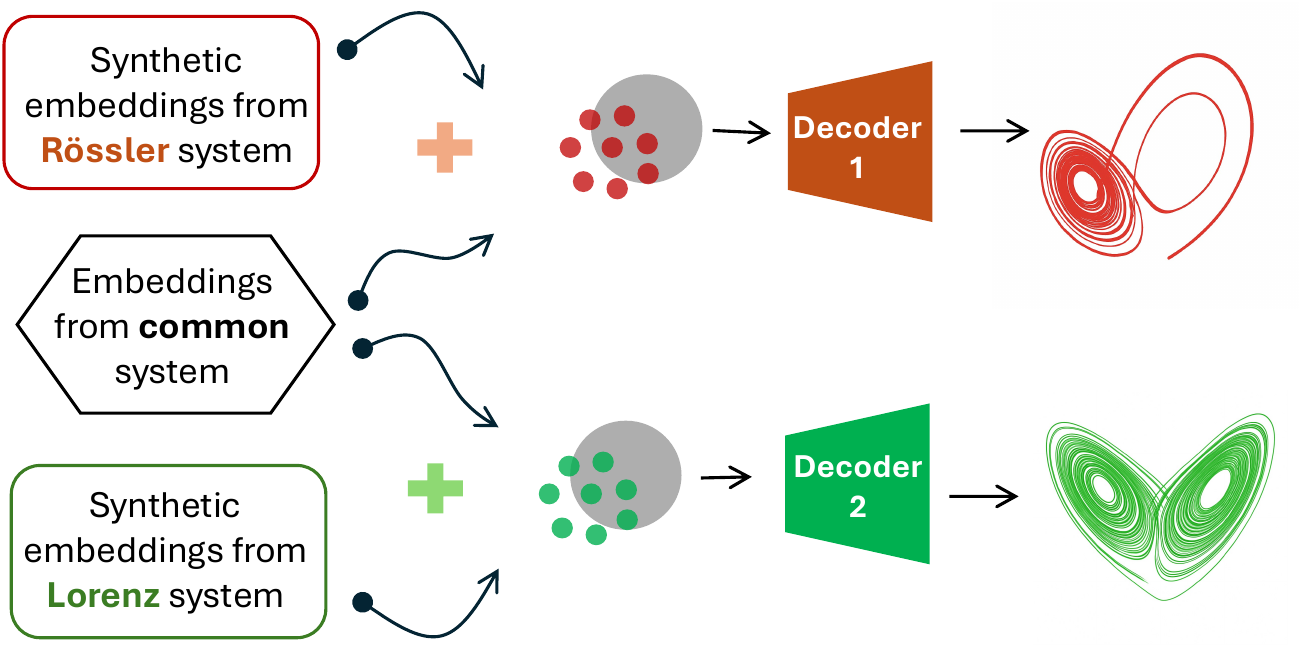}
    \caption{Generation of synthetic trajectories: A fixed embedding from the common latent space is combined with a set of system-specific embeddings generated by the SGM. The full latent vectors are decoded into physical trajectories using a trained autoencoder.}
    \label{fig:schema}
\end{figure}

\begin{figure}[h!]
    \centering
    \includegraphics[width=0.7\linewidth]{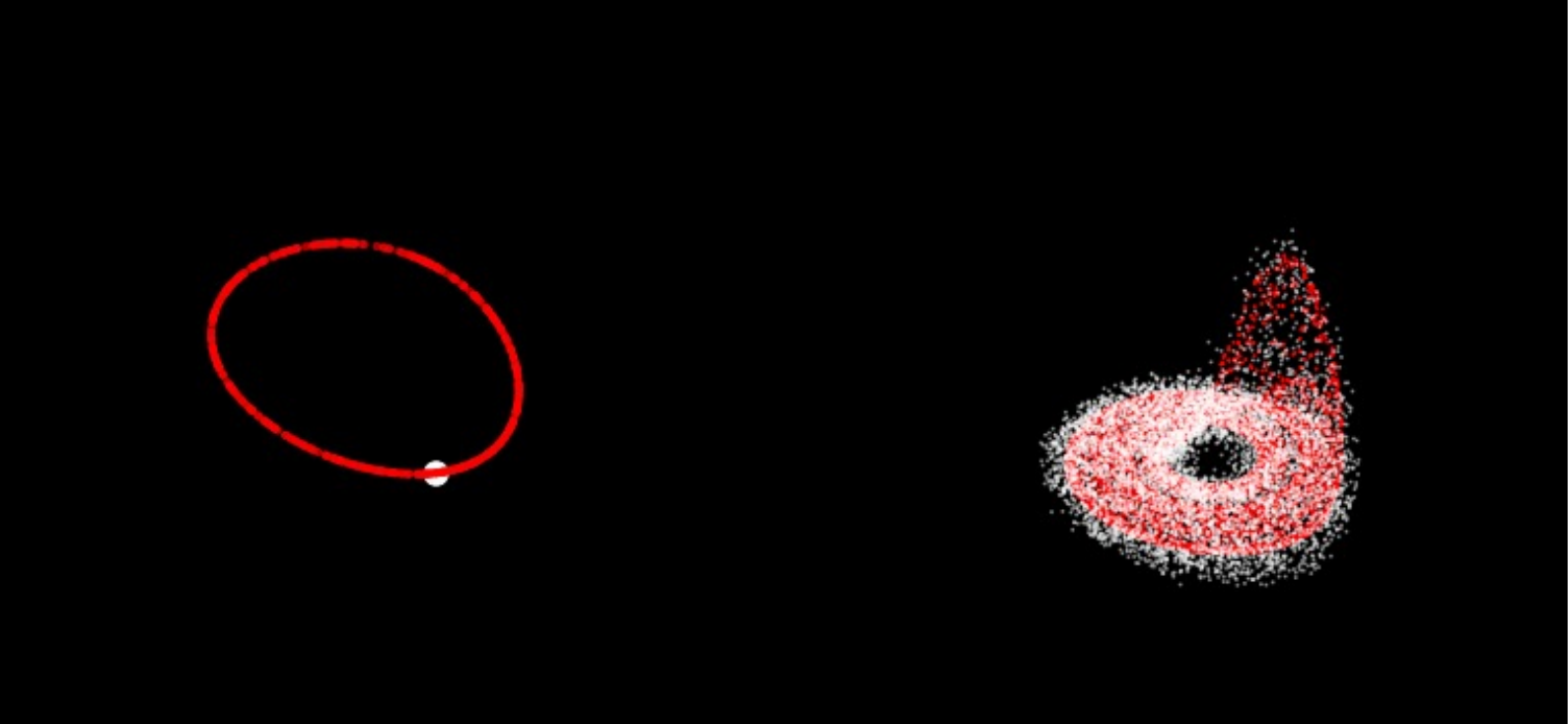}
    \caption{Synthetic reconstruction of the Rössler system. A fixed common embedding (left, white point) is paired with SGM-generated Rössler embeddings and decoded through the autoencoder to generate trajectories (right, white). The synthetic outputs preserve the structure of the original attractor.}
    \label{fig:SGM_Roessler}
\end{figure}

To translate the synthetic latent samples into physically meaningful trajectories, we fix a single embedding from the common latent space and pair it with SGM-generated embeddings corresponding to the Rössler modality. These concatenated representations are decoded via the first autoencoder (see Fig.~\ref{fig:schema}), yielding trajectories in the original physical space. The resulting outputs faithfully reproduce the geometry of the Rössler attractor (Fig.~\ref{fig:SGM_Roessler}), confirming that the generated latent embeddings are not only statistically consistent but also dynamically valid.

\begin{figure}[H]
    \centering
    \includegraphics[width=1.0\linewidth]{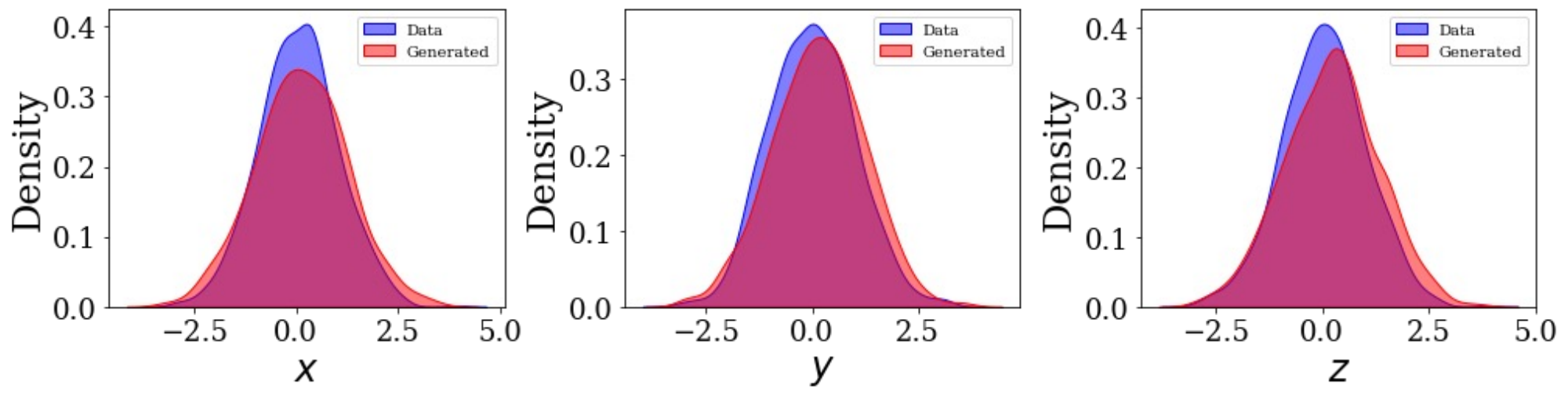}
    \caption{Lorenz embeddings: KDE plots comparing marginal distributions across latent dimensions. Moderate distributional mismatch is reflected in a KL divergence of 0.3595 and a Wasserstein distance of 0.1641.}
    \label{fig:SGM_dist_Lorenz}
\end{figure}

We apply the same procedure to the Lorenz system to evaluate generalization across modalities. Statistical comparison of the latent embeddings yields a KL divergence of 0.3595 and a Wasserstein distance of 0.1641, reflecting slightly better agreement compared to the Rössler case. KDE visualizations (Fig.~\ref{fig:SGM_dist_Lorenz}) confirm this alignment, although local deviations in low-density regions are evident.

\begin{figure}[H]
    \centering
    \includegraphics[width=0.7\linewidth]{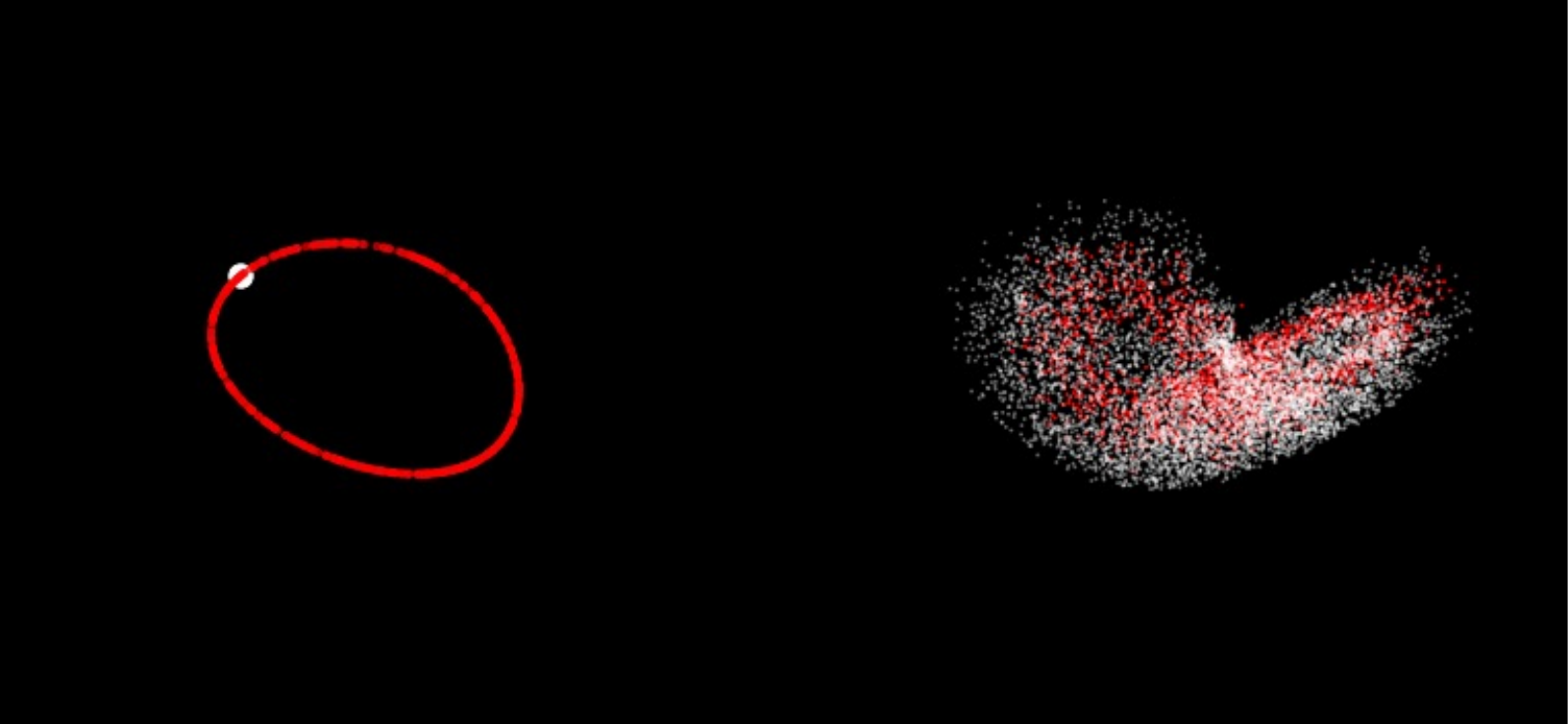}
    \caption{Reconstruction of the Lorenz attractor using synthetic latent codes. A fixed common latent vector (left, white point) is combined with SGM-generated Lorenz embeddings and decoded to yield physically consistent trajectories (right, white).}
    \label{fig:SGM_Lorenz}
\end{figure}

Decoding the Lorenz-specific embeddings, again using a fixed point in the common latent space, results in trajectories that closely replicate the classic butterfly-shaped structure of the Lorenz attractor (Fig.~\ref{fig:SGM_Lorenz}), demonstrating that the synthetic embeddings are both statistically and dynamically consistent.

\subsection{Synthetic embeddings using Probabilistic Learning on Manifolds}
\label{sec:PLOM_results}

We further evaluate generative modeling using Probabilistic Learning on Manifolds (PLoM), which produces samples concentrated on a low-dimensional representation of the latent space. A total of 120{,}000 synthetic embeddings were generated to match the distribution of the original Rössler and Lorenz latent representations.

\begin{figure}[H]
    \centering
    \includegraphics[width=1.0\linewidth]{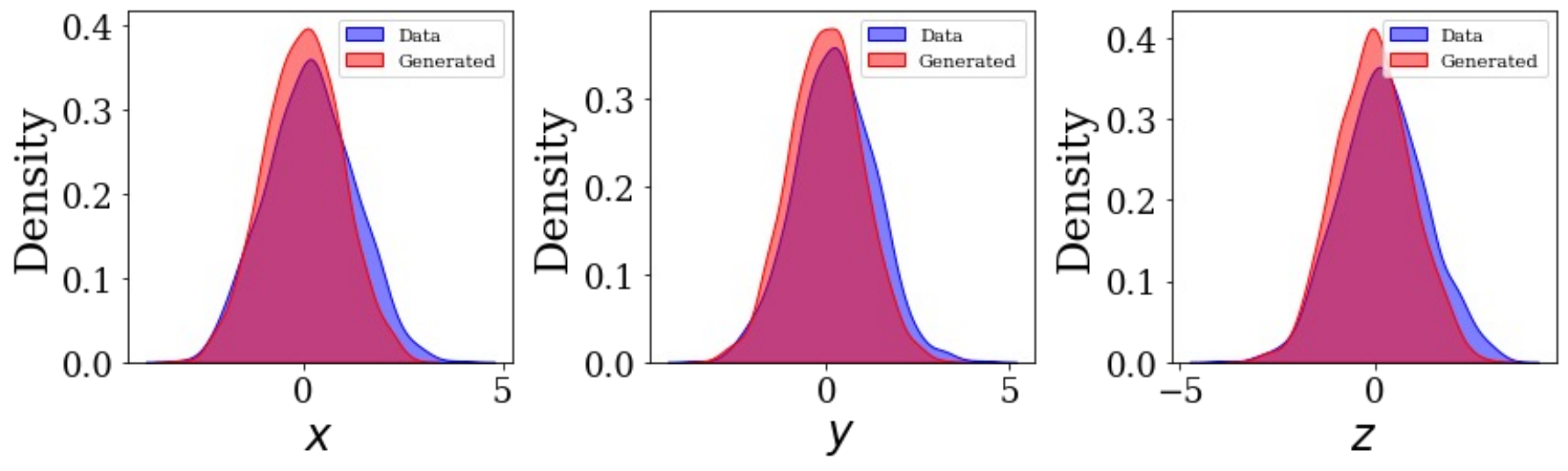}
    \caption{Rössler embeddings: Marginal distributions of latent coordinates for reference (blue) and PLoM-generated (red) samples. KL divergence of 0.5607 and Wasserstein distance of 0.2314 indicate moderate statistical divergence.}
    \label{fig:PLOM_dist_Rossler}
\end{figure}

As with the score-based model, we assess statistical fidelity using KL divergence and Wasserstein distance. For the Rössler system, the KL divergence between the reference and PLoM-generated embeddings is 0.5607, indicating a somewhat greater mismatch than observed with SGM. The mean Wasserstein distance, 0.2314, suggests moderate geometric alignment. Kernel density estimates of the marginal distributions reveal deviations in shape and location across dimensions (Fig.~\ref{fig:PLOM_dist_Rossler}).

\begin{figure}[H]
    \centering
    \includegraphics[width=0.7\linewidth]{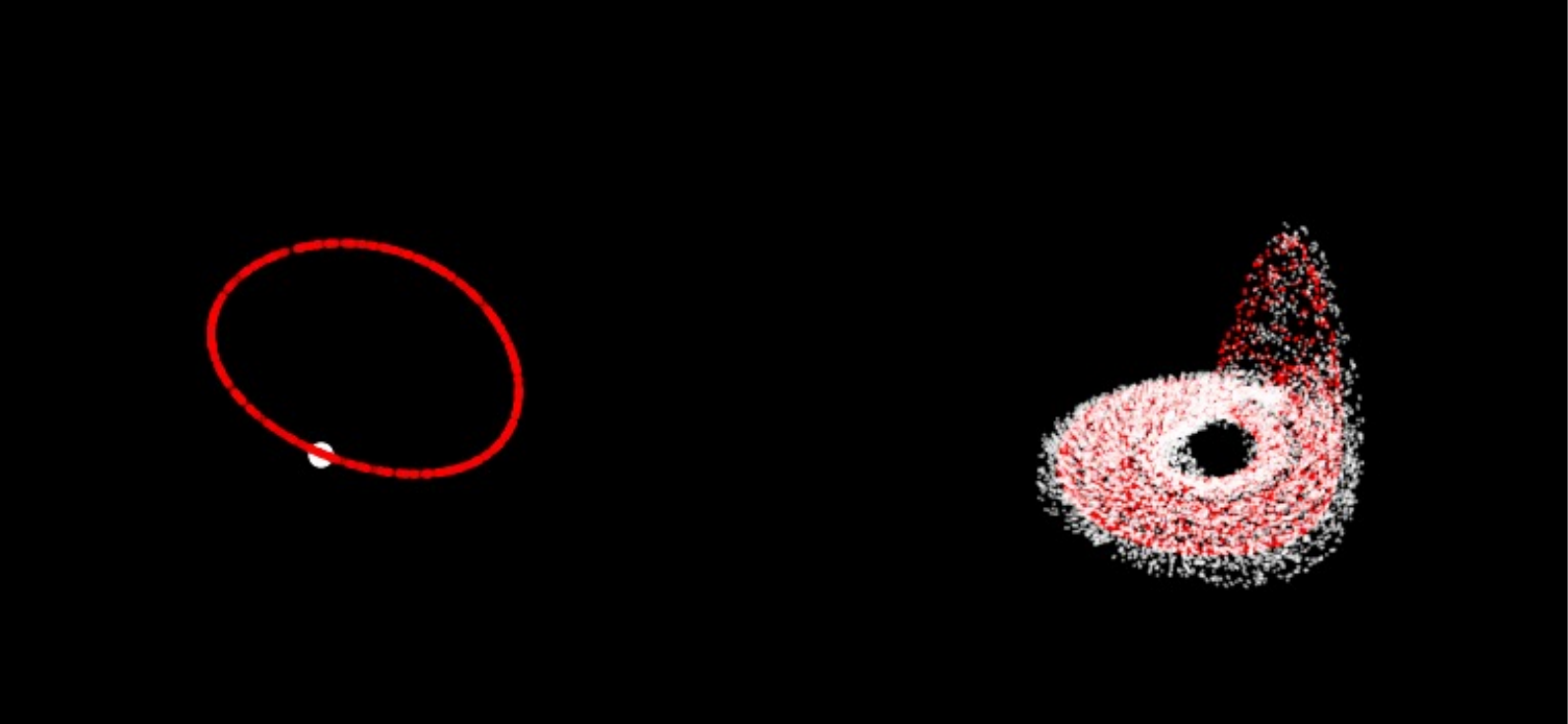}
    \caption{Reconstruction of the Rössler system using PLoM-generated embeddings. A fixed common latent vector (left, white point) is paired with synthetic Rössler-specific embeddings and decoded into the physical space (right, white), capturing the attractor structure.}
    \label{fig:PLOM_Roessler}
\end{figure}

To evaluate physical consistency, we decode synthetic latent codes by pairing them with a fixed embedding from the common latent space and passing them through the original autoencoder. The reconstructed trajectories closely follow the structure of the Rössler attractor (Fig.~\ref{fig:PLOM_Roessler}), confirming that the PLoM-generated embeddings, though distributionally less precise, are dynamically valid and physically meaningful.

\begin{figure}[H]
    \centering
    \includegraphics[width=1.0\linewidth]{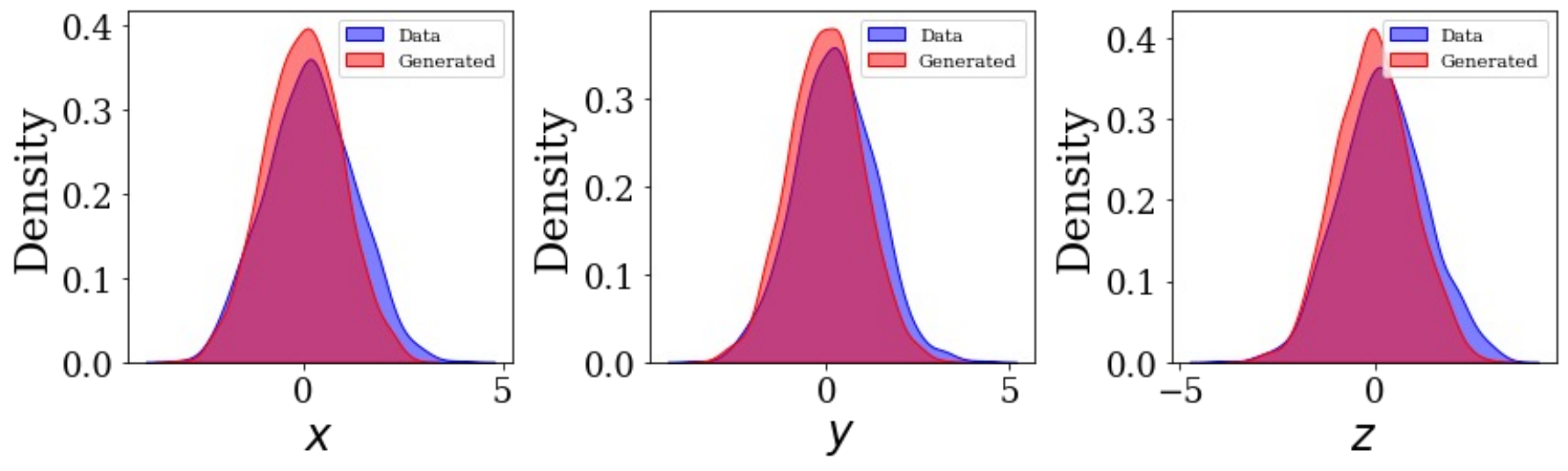}
    \caption{Lorenz embeddings: Marginal KDEs comparing reference (blue) and PLoM-generated (red) samples. A KL divergence of 0.4551 and Wasserstein distance of 0.1641 reflect moderate agreement.}
    \label{fig:PLOM_dist_Lorenz}
\end{figure}

We repeat this evaluation for the Lorenz system. The KL divergence between reference and generated embeddings is 0.4551, and the Wasserstein distance is 0.1641. These values suggest improved statistical alignment over the Rössler case. KDE plots across each latent dimension (Fig.~\ref{fig:PLOM_dist_Lorenz}) confirm this trend, though local differences in distribution persist.

\begin{figure}[H]
    \centering
    \includegraphics[width=0.7\linewidth]{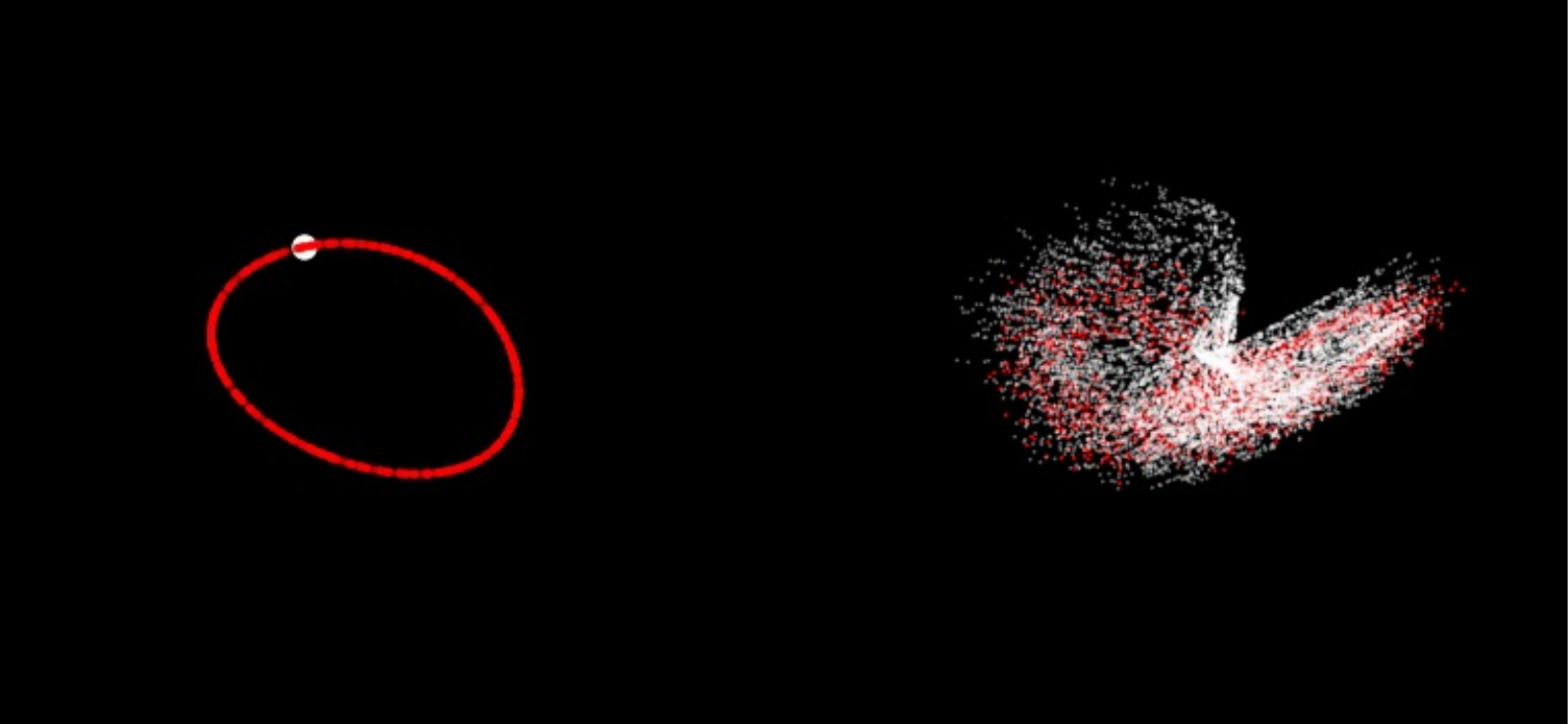}
    \caption{Lorenz attractor reconstructed from PLoM-generated latent codes. A fixed common latent embedding (left, white point) is combined with synthetic Lorenz-specific vectors and decoded into the physical space (right, white), recovering the system's characteristic structure.}
    \label{fig:PLOM_Lorenz}
\end{figure}

Once decoded, the synthetic latent vectors yield trajectories that accurately replicate the canonical butterfly pattern of the Lorenz attractor (Fig.~\ref{fig:PLOM_Lorenz}). These results validate the effectiveness of PLoM for generating latent embeddings that support physically plausible reconstructions, even when full probabilistic fidelity is not achieved.

\section{Discussion}\label{sec:discussion}

In this work, we proposed a structured autoencoder network architecture that, through training with a deliberately designed loss function, is capable of identifying `common' information between two (or, in principle, more) heterogeneous data sets, and subsequently of \textit{disentangling} and parametrizing information that is `uncommon', i.e. unique to each data set. We demonstrate this task in a dynamical system setting, where we are able to `recover' representations of underlying dynamical systems that are originally separated, but then `scrambled' by a set of rotations. We show that the framework can be used when observations come in the form of high-dimensional images, as well as when there is a time-lag between them.

The implemented bi-level optimization algorithm demonstrates that the identification of the common subspace is, in some sense, a separate and simpler problem than the parametrization of the uncommon subspaces. It is natural to first parametrize the common subspace, in order to be able to subsequently define independence from it. 

We note that the approach is completely unsupervised(!), given the key assumption of simultaneity in observations across the data sets, and that the correct latent space dimensions are known. The architecture is still amenable to a semi-supervised objective, when the common or some uncommon system have already been previously identified (see e.g. \cite{mishne2019259diffusionnet}). These can be prescribed as supervised objectives in the algorithm, which then would seek to `complete the picture' given this information (in analogy with the `gappy' data context we referenced above). This also allows our method to be combined with other algorithmic routines (e.g. Alternating Diffusion) to compute the common subspace in cases where autoencoders might struggle compared to spectral methods (e.g. high extrinsic curvature of the embedded data). While we use fully connected layers in our architecture implementation, orthogonality constraints can similarly be used for alternative architectures such as convolutional autoencoders \cite{masci2011stackedcae} to obtain disentangled latent representations.

Our definition of independence in this work is orthogonality of the subspaces on which the common and uncommon systems lie, in observation space. The algorithm attempts to find an embedding that makes these nonlinear subspaces mutually orthogonal. In that sense, it is purely geometric, in contrast to statistical disentanglement methods, such as nonlinear independent component analysis (NLICA) \cite{hyvarinen1999ICA}. For example, we note that the uncommon system's distribution need not be identical for every fixed point in the common system, it simply must lie in a (nearly) orthogonal subspace. Orthogonality is not affected by the boundaries of the observed data; conditional distributions are. Ultimately, depending on the problem at hand, the definition of independence matters. There may be problems where conformal and statistical independence coincide, though generically they do not. Such relations between geometric and statistical disentanglement warrant further study.

Regarding obstructions to conformal disentanglement, there are known results that limit the possibility of complete diagonalization of the metric tensor of a general manifold in dimensions 3 or 4 and above \cite{DeTurckDennisM.1984Eoed}. These do not directly apply to the present setting, since our metric is only block-diagonalized at particular points. Nevertheless, obstructions of a similar nature may hold. Fundamentally, the extent to which the data can be disentangled will be controlled by the structure of the Weyl tensor (generalized curvature) of the manifold which they lie on. Thus, it is, in some sense, controlled by the regularity of the embedding maps ($\Phi_u^{-1},\Phi_v^{-1}$ in the notation of \cref{sec:Problem_Statement}).

We state again that the setting easily generalizes to \textit{multiple} data sets (beyond just two) that may share common information.

An important feature of this work is that (a) for unscrambled, `clean' data, it allows us to learn a function between the different measurements of the common system by the two sensors: in effect, it constructs an {\em observer} of one sensor's measurement of the common system from the other. 
More importantly, however, it allows us, given a measurement from the first sensor, to (b) create a sampling of all possible second sensor measurements consistent with it - a ``level set" of what the second sensor could possibly be observing at that time instance. 

Interestingly, when one sensor measures a moment in time, and the second sensor measures the same variables at a later time, what is ``common" between the two sensors allows us to identify the dynamical system itself, thus establishing a type of correlational causality across components of the sensor measurements.

\section{Conclusion}\label{sec:conclusion}
We introduced a two-view autoencoder architecture that disentangles \emph{common} latent factors shared across measurements from \emph{uncommon} (view-specific) factors. A structured loss together with a bi-level optimization strategy promotes both accurate reconstruction and separation of latent subspaces, enabling interpretable latent swapping and cross-decoding experiments. This provides a fully unsupervised route to isolate shared structure from sensor- or modality-specific variability.

Building on this disentanglement, we proposed a \emph{curated generation} methodology: we learn a generative model directly on the learned latent coordinates (either the common or uncommon manifolds), draw samples from that model, and fuse sampled uncommon (or common) codes with real latent codes from data before decoding back to ambient space. This enables controlled synthesis that preserves desired shared attributes while varying targeted factors, and it offers a principled mechanism to generate curated ambient-space samples with specified features.

\subsection*{Data and Code Accessibility}
The synthetic data sets used in the computational examples, along with accompanying code will become publicly available upon publication. Regarding the bobble-head data set of \cref{exmp:bobbleheads} see the original publication \cite{lederman2014}.

\subsection*{Acknowledgements}
The authors are grateful to Dr. D. Sroczynski for aiding in the synthetic data set generation, and to Professor R. Lederman for giving us access to the bobble-head image data set originally featured in \cite{lederman2014}. The authors are thankful to Professors R. Lederman, R. Talmon, and R. Coifman for helpful discussions and suggestions regarding this project. EDK was funded by the Luxembourg National Research Fund (FNR), grant reference 16758846. The work of D.G. and I.G.K. was partially supported by the US National Science Foundation and the US Department of Energy; 

\newpage

\bibliographystyle{unsrt}
\bibliography{merged_bibliography}

\newpage
\appendix
\section{Data and Network Architectures}\label{app:net_data}

\subsection{Data Generation}

\textbf{\cref{exmp:Torus}}
The data consists of samples of trajectories of three harmonic oscillations. The dynamics, i.e. the angle as a function of time $\theta(t)$ have the form:
\begin{align*}
    \theta_i(t) = 2\pi \omega_i t
\end{align*}
where the common and uncommon systems are characterized by
\begin{align*}
    \mcU:\omega_1=\frac{\pi}{2}\qc
    \mcV:\omega_2=\frac{\pi}{2\sqrt{2}}\qc
    \mcC:\omega_3=\frac{1}{\sqrt{2}}
\end{align*}
Note that the ratios between each system's frequencies are irrational, so as to avoid resonances. We sample a total of 3000 temporally equidistant points for each system. We `scramble' the data using an invertible linear transformation, such that individual coordinates do not correspond to a single particular system for each sensor.

\textbf{\cref{exmp:RL}}
\label{app:data+generation}
The data consists of samples of trajectories for the following:

$\mathcal{U}$ is sampled from a R\"{o}ssler attractor:
\begin{align}
    \begin{split}
        \dot{x}&=-y-z\\
        \dot{y}&=x+\alpha y\\
        \dot{z}&=b+z(x-c)
    \end{split}
\end{align}
with $\alpha=b=0.2$ and $c=5.7$.

$\mathcal{V}$ is sampled from a Lorenz attractor:
\begin{align}
    \begin{split}
        \dot{x}&= \sigma(y-x)\\
        \dot{y}&=x(\rho-z)-y\\
        \dot{z}&=xy-\beta z
    \end{split}
\end{align}
with $\sigma=10$, $\beta=8/3$, $\rho=28$.

$\mathcal{C}$ is sampled from a nonlinear limit cycle:
\begin{align}
    \begin{split}
        \dot{x}&=\alpha_1(1-x-y)-\gamma_1x-xy(1-x-y)^2\\
        \dot{y}&=\alpha_2(1-x-y)-\gamma_2y-xy(1-x-y)^2
    \end{split}
\end{align}
with $\alpha_1=0.016$, $\gamma_1=0.001$, $\alpha_2=0.0278$, $\gamma_2=0.002$.

We sample a total of 3000 temporally equidistant points for the data set, using accurate numerical simulation. We `scramble' the data using an invertible linear transformation, such that individual coordinates do not correspond to a single particular system for each sensor.

\textbf{\cref{exmp:bobbleheads}} The original data consists of colored RGB images of $320\times240$ pixels. There are 8100 images for each sensor (16200 total). Each bobble-head rotates circularly in one direction with an unknown frequency, while the frame rate is also unknown.

We randomly select $2500$ pairs of snapshots as a training set, $700$ as a validation set, and $1000$ as a test set before optimization. These images are converted to gray-scale (single value per pixel), and compressed into square images with $240\times240$ pixels. Subsequently, they are projected onto their first $60$ principal components, which are what the autoencoder architecture trains on. The principal components are computed per sensor, on the full set of $4200$ randomly selected images.

\textbf{\cref{sec:causal}}
The data consists of samples of trajectories for dynamical systems. $\mcC$ is a common limit cycle (as in \cref{exmp:Torus}) which sensor 1 observes  at time $t$, and sensor 2 observes at time $t+\Delta t$, where $\Delta t=200$. The uncommon system $\mcU$ observed by the second sensor is a R\"{o}ssler attractor (as in \cref{exmp:RL}). Note that sensor 1 does not observe any other system in this example.

We collect 2800 samples in time (for each system), and use a train/validation/test split of 54\%, 36\%, 10\% respectively. The train and validation errors after the first optimization step are \num{1.2e-4} and $\num{1.5e-4}$, for the run used to produce \cref{fig:causal}.

\begin{table}[ht]
    \centering
    \begin{tabular}{ccc}
        & $\min$ & $\max$\\ \cline{2-3}
        \cref{exmp:Torus} & -2.20 & 2.20\\ \cline{2-3}
        \cref{exmp:RL} &-3.66 &7.73\\ \cline{2-3}
        \cref{exmp:bobbleheads} &-8.02&10.83\\ \cline{2-3}
        \cref{sec:causal} & -2.08 & 8.11
    \end{tabular}
    \caption{Minimum and maximum values of data both sensors after pre-processing, used to train the autoencoder weights.}
    \label{tab:data_min_max}
\end{table}

We include nominal ranges of the data for each example after pre-processing in \cref{tab:data_min_max}. These were used to train the autoencoder architecture in each case.

\subsection{Architecture Details}

Our neural networks are implemented in \texttt{Python} using \texttt{PyTorch} \cite{NEURIPS2019_9015}.

\textbf{\cref{exmp:Torus}} All encoders and decoders consist of 7 fully connected linear layers the first 5 of which are composed with a $\tanh$ activation function. Each encoder layer has a width of 10 nodes and each decoder has a width of 20. Each encoder $\mfe_{u}^c,\mfe_{v}^c,\mfe_1^u,\mfe_2^u$ maps into two-dimensional latent subspaces, while decoder $\mfd_1, \mfd_2$ map the corresponding 4-dimensional latent spaces back into a Euclidean space of the same dimension $(\R^4)$.

\textbf{\cref{exmp:RL}} All encoders and decoders consist of 7 fully connected linear layers the first 5 of which are composed with a $\tanh$ activation function. Each layer has a width of 30. Each common encoder $\mfe_{u}^c,\mfe_{v}^c$ maps to a two-dimensional common latent subspace, and each uncommon encoder $\mfe_{u}^c,\mfe_{v}^c$ maps to separate three-dimensional uncommon latent subspaces. Each decoder $\mfd_1, \mfd_2$ map the corresponding latent spaces back into a Euclidean space of the same dimension $(\R^5)$. A sketch with the corresponding dimensions can be found in \cref{fig:architecture}.

\textbf{\cref{exmp:bobbleheads}} All encoders consist of 7 fully connected linear layers the first 5 of which are composed with the $\tanh$ activation function. Each layer has width 40. Each encoder $\mfe_{u}^c,\mfe_{v}^c,\mfe_1^u,\mfe_2^u$ maps into two-dimensional latent subspaces. Each decoder $\mfd_1, \mfd_2$ has the same fully-connected structure as the encoder networks with a width of 80 nodes within each layer. They map the corresponding 4-dimensional latent spaces back into a Euclidean space $\R^{60}$, the truncated principal component representation of the images.

\textbf{\cref{sec:causal}}
\section{Optimization Algorithm}\label{app:algs}
\cref{alg:Common_Uncommon_CAE} presents a more detailed sketch of the optimization scheme proposed in \cref{sec:opti}. In practice, we use Adam \cite{kingma2014adam} instead of (Stochastic) Gradient Descent. We also note that we partition given data sets into train (used by the algorithm to evaluate and train on) validation (used by the algorithm during training to check performance on `unseen' data) and test (only used after successful convergence to evaluate the performance of the algorithm) sets. This is common practice in network training and serves to avoid overfitting, but not represented in \cref{alg:Common_Uncommon_CAE} for the sake of clarity. The respective errors in each computational example are reported in \cref{sec:examples}.

\begin{algorithm}
	\caption{CAE - Common Orthogonal Latent Decomposition}
	\label{alg:Common_Uncommon_CAE}
	\KwData{Data sets ($S_2$, $S_1$)}
        \KwParams{initialized CAE architecture  and weights ($\mfe_1^c,\mfe_1^u,\mfe_2^c\mfe_2^u,\mfd_1,\mfd_2$). Learning rate $\eta$, error tolerances $\epsilon_1,\epsilon_2$ are initialized to be small positive constants}
	\KwResult{`Orthogonally disentangled' common - uncommon latent representation of data}
\vspace{.1cm}
\hrule
\vspace{.1cm}
\textbf{1st Level} - Common subspace identification
\vspace{.1cm}

 \While{$\mathcal{L}_\text{CAE}\geq \epsilon_1$}{
		  // encoding and decoding forward pass
		\begin{gather}
                \qty{(\hat{\vb{u}}_i,\hat{\vb{c}}_{ui})}_{i=1}^N=\qty{(\mfe_1^c(\vb{s}_{ui}),\mfe_1^u(\vb{s}_{ui}))}_{i=1}^N\\
                \qty{(\hat{\vb{v}}_i,\hat{\vb{c}}_{vi})}_{i=1}^N=\qty{(\mfe_2^c(\vb{s}_{vi}),\mfe_2^u(\vb{s}_{vi}))}_{i=1}^N
            \end{gather}
            \begin{gather}
                \qty{\hat{\vb{s}}_{ui}}_{i=1}^N=\qty{\mfd_1(\hat{\vb{u}}_i,\hat{\vb{c}}_{ui})}_{i=1}^N\\
                \qty{\hat{\vb{s}}_{vi}}_{i=1}^N=\qty{\mfd_2(\hat{\vb{v}}_i,\hat{\vb{c}}_{vi})}_{i=1}^N
		\end{gather}
		
            // compute reconstruction loss and common subspace loss
		\begin{equation}
			\mathcal{L}_\text{CAE}=\underbrace{\frac{1}{N}\sum_{i=1}^N\norm{\vb{s}_{ui}-\hat{\vb{s}}_{ui}}_2^2+\norm{\vb{s}_{vi}-\hat{\vb{s}}_{vi}}_2^2}_{\qq{Reconstruction}}+\underbrace{\frac{1}{N}\sum_{i=1}^N\norm{\hat{\vb{c}}_{ui}-\hat{\vb{c}}_{vi}}_2^2}_{\qq{Common Matching}}
		\end{equation}
		// perform backward pass
		\begin{align}
			w_{\mathfrak{e}_u^u}\mathrel{-}=\eta\grad_{w_{\mathfrak{e}_u^u}}\mathcal{L}_\text{CAE}\qc
                w_{\mathfrak{e}_u^c}\mathrel{-}=\eta\grad_{w_{\mathfrak{e}_u^c}}\mathcal{L}_\text{CAE}\\
                w_{\mathfrak{e}_v^u}\mathrel{-}=\eta\grad_{w_{\mathfrak{e}_v^u}}\mathcal{L}_\text{CAE}\qc
                w_{\mathfrak{e}_v^c}\mathrel{-}=\eta\grad_{w_{\mathfrak{e}_v^c}}\mathcal{L}_\text{CAE}\\
			w_{\mathfrak{d}_u}\mathrel{-}=\eta\grad_{w_{\mathfrak{d}_u}}\mathcal{L}_\text{CAE}\qc
                w_{\mathfrak{d}_v}\mathrel{-}=\eta\grad_{w_{\mathfrak{d}_v}}\mathcal{L}_\text{CAE}
		\end{align}
	}
 \textbf{2nd Level} - Uncommon subspace identification
 \vspace{.1cm}
 
 \While{$\mathcal{L}_\text{CAE}\geq \epsilon_2$}{
 encoding and decoding forward pass (same as in previous level)

 // compute reconstruction and orthogonality loss
 \begin{multline}
     \mathcal{L}_\text{CAE}=\underbrace{\frac{1}{N}\sum_{i=1}^N\norm{\vb{s}_{ui}-\hat{\vb{s}}_{ui}}_2^2+\norm{\vb{s}_{vi}-\hat{\vb{s}}_{vi}}_2^2}_{\qq{Reconstruction}}\\+\underbrace{\frac{1}{N}\sum_{i=1}^N\sum_{j, k}\norm{\expval{\grad(\hat{\vb{c}}_{ui})_j,\grad(\hat{\vb{u}}_i)_k}}^2_2+\frac{1}{N}\sum_{i=1}^N\sum_{j,l}\norm{\expval{\grad(\hat{\vb{c}}_{vi})_j,\grad(\hat{\vb{v}}_i)_l}}^2_2}_{\qq{Common-Uncommon Orthogonality}}
 \end{multline}
 // perform backward pass
		\begin{align}
			w_{\mathfrak{e}_u^u}\mathrel{-}=\eta\grad_{w_{\mathfrak{e}_u^u}}\mathcal{L}_\text{CAE}\qc
                w_{\mathfrak{e}_v^u}\mathrel{-}=\eta\grad_{w_{\mathfrak{e}_v^u}}\mathcal{L}_\text{CAE}\\
			w_{\mathfrak{d}_u}\mathrel{-}=\eta\grad_{w_{\mathfrak{d}_u}}\mathcal{L}_\text{CAE}\qc
                w_{\mathfrak{d}_v}\mathrel{-}=\eta\grad_{w_{\mathfrak{d}_v}}\mathcal{L}_\text{CAE}
		\end{align}
}
\end{algorithm}
In \cref{alg:Common_Uncommon_CAE}, index $i$ iterates over the samples of the data set, while indices $j,k,l$ iterate over the dimensions ($n_c,n_u,n_v$) of the common and uncommon subspaces respectively.

\subsection{Alternative Architectures}\label{sec:alt_architecture}

In the original architecture, we minimize the `common' constraint $\mcL_\text{common}$ to bias the network towards finding the desired latent representation for the common variables. It is known, however, that minimizing multiple objectives in a loss function using gradient-descent-type algorithms can be difficult when the proper weighting of each term is unknown.

While the proposed implementation and algorithm in \cref{sec:methods} is sufficient for the dynamical system computational examples, we observed this process not to be as stable in the case of \cref{exmp:bobbleheads}, where the ambient dimension was significantly larger. In particular, minimizing $\mcL_\text{common}$ in this case resulted in the AE  converging to `noise' instead of any structured latent representation of the common variable. Reconstructions of the images featured an `averaged' image for the bulldog, over its entire circular trajectory.

\begin{figure}[ht]
	\centering
	\includegraphics{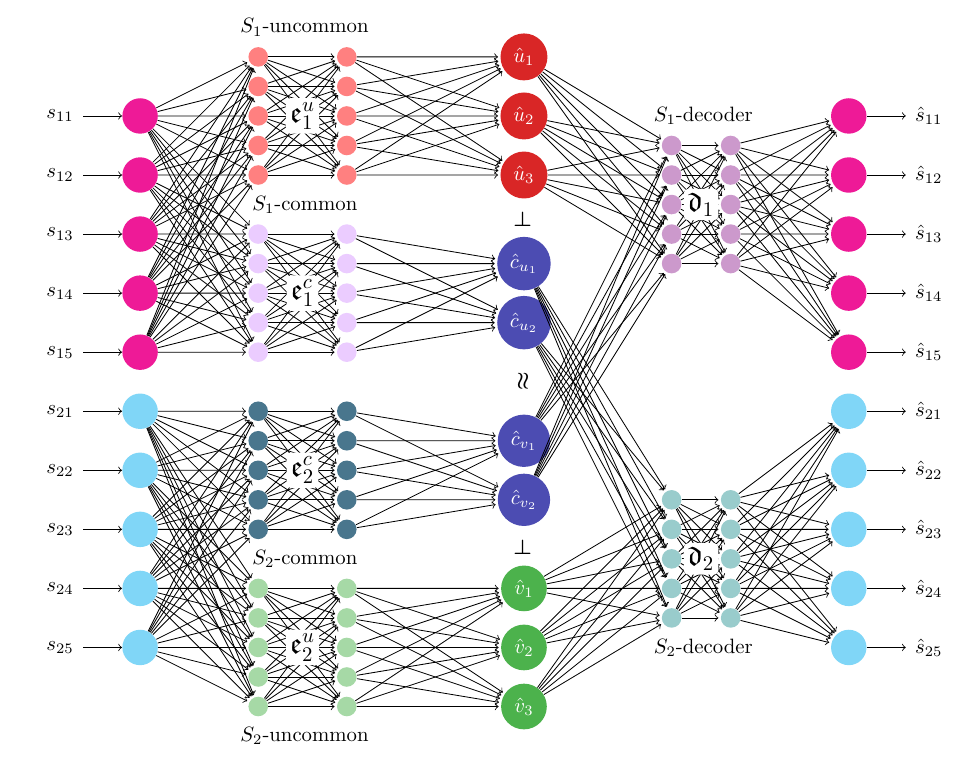}
	\caption{The `Twisted' architecture is a variant of the original proposed architecture, where we bias the network to find a good common latent space by asking the reconstruction of each sensor to be influenced by the input to both sensors. This is achieved by changing the input to each decoder to use the `common' information from the \textit{other} sensor's encoder.}
	\label{fig:twisted_architecture}
\end{figure}

To remedy this issue, we propose the `Twisted' variant of the architecture represented in \cref{fig:twisted_architecture}. There, we `force' the decoder for each sensor to view information from the encodings of both sensors. The only information from the first sensor relevant to the second one is, of course, related to the common variables, and \textit{vice versa}. This architecture is capable of identifying the common submanifold (up to diffeomorphism) without minimizing an additional constraint.

We empirically found this architecture to be more robust to this convergence issue during training, and used it to produce the results in \cref{exmp:bobbleheads}. After identifying the common submanifold, the inputs to the decoders can be switched to the original configuration of \cref{fig:architecture} before disentangling via the second step of \cref{alg:Common_Uncommon_CAE}.

\end{document}